\documentclass[sigconf, nonacm]{acmart}

\AtBeginDocument{%
  }

\usepackage{makecell}
\usepackage{xcolor}
\usepackage{todonotes}
\usepackage{threeparttable}

\usepackage{caption}
\usepackage{subcaption}
\usepackage{multirow}
\usepackage{rotating} 

\setcopyright{licensedusgovmixed}\acmConference[SIGSPATIAL '24]{The 32nd ACM International Conference on Advances in Geographic Information Systems}{October 29-November 1, 2024}{Atlanta, GA, USA}
\acmBooktitle{The 32nd ACM International Conference on Advances in Geographic Information Systems (SIGSPATIAL '24), October 29-November 1, 2024, Atlanta, GA, USA}
\acmDOI{10.1145/3678717.3691292}
\acmISBN{979-8-4007-1107-7/24/10}

\begin{document}
\title[OReole-FM: successes and challenges toward billion-parameter FMs for high-resolution satellite imagery]{OReole-FM: successes and challenges toward billion-parameter foundation models for high-resolution satellite imagery}

\author{Philipe Dias}
\author{Aristeidis Tsaris}
\email{ambroziodiap@ornl.gov}
\affiliation{%
  \institution{Oak Ridge National Laboratory}
  \city{Oak Ridge}
  \state{Tennessee}
  \country{USA}
}
\author{Jordan Bowman}
\author{Abhishek Potnis}
\author{Jacob Arndt}
\affiliation{%
  \institution{Oak Ridge National Laboratory}
  \city{Oak Ridge}
  \state{Tennessee}
  \country{USA}
}
\author{H. Lexie Yang}
\author{Dalton Lunga}
\email{lungadd@ornl.gov}
\affiliation{%
  \institution{Oak Ridge National Laboratory}
  \city{Oak Ridge}
  \state{Tennessee}
  \country{USA}
}



\renewcommand{\shortauthors}{Dias, P.; Tsaris, A.; Bowman, J.; Potnis, A.; Arndt, J.; Yang, H.L.; Lunga, D.}



\begin{abstract}

While the pretraining of Foundation Models (FMs) for remote sensing (RS) imagery is on the rise, models remain restricted to a few hundred million parameters. Scaling models to billions of parameters has been shown to yield unprecedented benefits including emergent abilities, but requires data scaling and computing resources typically not available outside industry R\&D labs. In this work, we pair high-performance computing resources including Frontier supercomputer, America's first exascale system, and high-resolution optical RS data to pretrain billion-scale FMs. Our study assesses performance of different pretrained variants of vision Transformers across image classification, semantic segmentation and object detection benchmarks, which highlight the importance of data scaling for effective model scaling. Moreover, we discuss construction of a novel TIU pretraining dataset, model initialization, with data and pretrained models intended for public release. By discussing technical challenges and details often lacking in the related literature, this work is intended to offer best practices to the geospatial community toward efficient training and benchmarking of larger FMs. 
\end{abstract}    
\maketitle
\let\thefootnote\relax\footnotetext{This is the author's full version of the work, published in ACM SIGSPATIAL 2024 as a short paper. Not for redistribution. The definitive Version of Record was published in https://doi.org/10.1145/3678717.3691292. This manuscript has been authored by UT-Battelle, LLC, under contract DE-AC05-00OR22725 with the US Department of Energy (DOE). The US government retains and the publisher, by accepting the article for publication, acknowledges that the US government retains a nonexclusive, paid-up, irrevocable, worldwide license to publish or reproduce the published form of this manuscript, or allow others to do so, for US government purposes. DOE will provide public access to these results of federally sponsored research in accordance with the DOE Public Access Plan (http://energy.gov/downloads/doe-public-access-plan). This research used resources of the Oak Ridge Leadership Computing Facility, which is a DOE Office of Science User Facility supported under Contract DE-AC05-00OR22725.}
\section{Introduction}

Remote sensing (RS) imagery interpretation remains pivotal to understanding of the Earth surface, with Earth Observation (EO) applications including mapping of built environments~\cite{arndt2021urbanstructural}, disaster management~\cite{laverdiere2020rapid}, and gravity mapping~\cite{yang2021agu}. While deep neural networks (DNN) and computer vision have enable extraordinary progress in interpretation of imagery, most DNNs are task-specific, with limited generalization to out-of-distribution data, and rely on the onerous availability of large quantities of manually annotated data samples.

Recently, Foundation Models (FMs) have emerged as a breakthrough with potential to address limitations of neural networks in terms of task-specificity, poor generalization to out-of-distribution data, and reliance on labeled data. FMs can be defined as large models ($>10^8$ parameters) usually trained via self-supervised learning (SSL) on vast data volumes, and are characterized by extraordinary transfer learning capabilities on a wide range of downstream tasks. 

\begin{table}[b]
\centering
\begin{threeparttable}
  \caption{A comparison of model sizes for current state-of-the-art foundation models for Earth observation.}
  \label{tab:papers_sizes}
  \begin{tabular}{lcc}
    \toprule
    \textbf{Model} & \textbf{Backbone} & \textbf{Parameters} \\
    \midrule
    GASSL \cite{ayush2021geography} & ResNet-50 & $\sim25M$ \\
    SeCo \cite{Manas2021ICCV}& ResNet-50 & $\sim25M$ \\
    SatMAE \cite{cong2022satmae} & ViT-L & $\sim307M$ \\
    ScaleMAE \cite{reed2023scalemae} & ViT-L & $\sim307M$ \\
    RVSA \cite{wang2022rvsa} & ViT/ViTAE*-B & $\sim110M$ \\
    RingMo \cite{sun2022ringmo} & Swin/ViT-B & $\sim100M$ \\
    Prithvi \cite{jakubik2023foundation} & ViT-L & $\sim307M$  \\
    Satlas \cite{Bastani2022SatlasPretrainAL} & Swin-B & $\sim100M$  \\
    GFM \cite{mendieta2023towards}& Swin-B & $\sim100M$  \\
    USat \cite{irvin2023usat}& ViT-L & $\sim307M$  \\
    MTP \cite{wang2024mtp} & ViT-L/InternImage-XL & $\sim307/335M$ \\
    SkySense\cite{Guo2023SkySenseAM}\tnote{1} & Swin-H & $\sim654M$  \\
    \midrule
    \textsc{OReole-MR/-HR} & ViT-e (3B) & $\sim 3B$ \\
  \bottomrule
\end{tabular}
\begin{tablenotes}\footnotesize
\item[1] SkySense includes separate encoders for high-resolution (Swin-H) and medium-resolution (ViT-L, 302M) optical imagery, SAR (ViT-L, 302M). Combined, all modules composing the model add up to 2B parameters; 
\end{tablenotes}
\end{threeparttable}
\end{table}

A growing body of self-supervised studies is emerging in the remote sensing community, and Vision Transformers(ViTs) \cite{dosovitskiy2020vit} are by far dominating the large scale interpretation of remote sensing imagery. Table \ref{tab:papers_sizes} provides a summary on the state-of-the-art works pairing self-supervised learning and remote sensing data, along with model size configurations. Compared to the $100B+$ parameters powering modern LLMs, current FMs for RS are only up to few hundred million parameters. LLMs such as LLAMA (65B) \cite{touvron2023llama}, GPT-3 (175B) \cite{brown2020gpt3} and PaLM (540B) \cite{chowdhery2023palm}, while ViT-based structures have been extended for up to $20B+$ parameters \cite{dehghani2023scaling}. 
Importantly, studies in both NLP and computer vision have shown how scaling model capacity enables highly capable and generalizable FMs, including enabling the so-called emergent abilities: i.e., capabilities that are not present in smaller-scale models but are present in large-scale models, and cannot be predicted by simply extrapolating the performance improvements on smaller-scale models. Examples include capabilities that the model is not explicitly programmed or optimized for, such as  NLP models trained for predicting the next word in a sentence developing abilities to perform translation between languages or multiplying numbers \cite{wei2022emergent}.

However, scaling models to billions of parameters commands significant computing resources that have been typically restricted to on-premise high-performance computing (HPC) resources not available to researchers outside industry R\&D labs. The ViT-22B described in \cite{dehghani2023scaling} by Google Research was trained on 1024 TPU V4 chips, while the training of LLaMA by Meta AI on its 65B-parameter configuration was performed using 2048 NVIDIA A100 (80GB) GPUs \cite{touvron2023llama}. In addition to the accessibility to such computing resources, the expertise and best practices on how to leverage them effectively to train FMs remains limited to few established organizations. 

In this paper, we pretrain and evaluate billion-scale FMs for RS data analysis by leveraging the exascale Frontier Supercomputer \cite{FrontierWebsite}- the first exascale high performance computer. The study represents an initial step part of a broader agenda targeting best practices for training FMs, construction of pretraining datasets, benchmarking, and increasing the practical use of FMs for RS. As main contributions, we introduce:
\begin{itemize}
    \item The \textsc{OReole}\footnote{OReole is a play of words. OR alludes to the Oak Ridge National Laboratory, where this research is conducted. We believe birds are a great metaphor for such models for Earth Observation ("eyes from above"), and found the oriole family to be a perfect fit: orioles are songbirds found in many regions throughout the world, and their colorful feathers also offer a metaphor for the multimodality intended for this family of models. We then slightly adapt the name by including `EO`.} family of FMs for EO. The \textsc{OReole-MR} variant is pretrained on the MillionAID dataset of aerial RGB remote sensing images. Meanwhile, the \textsc{OReole-HR} is trained on very high-resolution BGR+NIR satellite imagery, targeting applications such as building footprint mapping  \cite{yang2018building, lunga2021resflow} and damage assessment \cite{dias2024bdicond} that are critical for, e.g., the development of LandScan's population maps \cite{bhaduri2007landscan, sims2023landscan}.

    \item ViT-based models scaled into the billion-parameter range, discussing successes, recommended HPC practices, as well as challenges for pretraining and evaluation; 

    \item \textbf{TIU}, a high-resolution RS imagery pretraining dataset that considers geographic, temporal, and image collection characteristics to benefit dataset diversity and representativeness; 
    
    \item A transfer learning technique for initialization of a 4-band model using pretrained weights from a 3-band (RGB) model, improving pretraining speed and downstream performance;
    
    \item All models, codes, and datasets in this work are intended for public release. Once approved, together with supplementary materials they will be published to \url{geoai.ornl.gov}.
\end{itemize}
\label{sec:intro}
\section{Related Works}
\label{sec:relworks}

\noindent\textbf{Key ingredients for developing FMs.} By leveraging surrogate tasks as sources of supervision, self-supervised learning (SSL) unlocks the potential of learning from unprecedentedly large unlabeled datasets for which manual labeling would be unfeasible, thus underpining the success of FMs. Masked modeling and contrastive learning (CL) have been the two prevailing approaches for SSL of computer vision models. Generative modeling schemes such as masked language modelling were key enablers for, e.g., GPT and BERT models \cite{radford2018gpt,devlin2018bert}. They have been analogously explored for computer vision in the form of masked image modeling (MIM) \cite{he2022masked}, where the model is tasked to reconstruct pixels of masked image patches based on the remaining visible image patches. Alternatively, CL schemes task models to maximize feature similarity for correlated image views constructed through augmentations \cite{chen2020simclr}. 

To enable scaling models to billions of parameters, in addition to SSL three key ingredients have been needed: scalable architectures, large data volumes, and computational resources. Transformer-based architectures \cite{vaswani2017transformers} have been widely used across domains, including Vision Transformers (ViT) \cite{dosovitskiy2020vit} for scalable image analysis models \cite{zhai2021scalingvit, dehghani2023scaling}. As discussed in \cite{xie2023data}, large models can easily overfit if data volumes are not scaled accordingly. Combined to the need for long pretraining, this implies the need for leveraging HPC resources for both fitting larger models in GPU memory, as well as enabling distributed data parallelism. 

\noindent\textbf{Foundation models in RS.} Transformer-based architectures have been the predominant strategy in works discussing FMs for RS. Masked Autoencoder (MAE) has been predominant for SSL, including SatMAE \cite{cong2022satmae}, Ringmo \cite{sun2022ringmo}, Prithvi \cite{jakubik2023foundation}, \cite{Hong2023SpectralGPTSR}, SATLASNet \cite{Bastani2022SatlasPretrainAL} and most recently the billion-scale foundation model (BFM) for remote sensing images \cite{cha2023billion}. Works including SeCo \cite{Manas2021ICCV}, GASSL \cite{ayush2021geography}, and Skysense \cite{Guo2023SkySenseAM} employ Contrastive Learning (CL) instead. In contrast to natural images, RS imagery allows leveraging image acquisitions covering the same location, but from different timestamps and sensors as augmentations for CL. Both GASSL and SeCo are designed to exploit spatially aligned images and temporal information to obtain seasonal positive pairs of images at different points in time for training seasonal contrasting learning objectives. In contrast, Skysense expoits features in a multi-granularity scheme to learn representations across different modal and spatial granularities.

Model’s architecture and training regime that exploit the characteristic features intrinsic in remote sensing imagery have been introduced, including image content \cite{wang2022rvsa}, spectra \cite{Hong2023SpectralGPTSR}, multi-scale features \cite{reed2023scalemae}, temporal characteristics \cite{cong2022satmae,jakubik2023foundation}. As summarized in Table \ref{tab:papers_sizes}, with the exception of Skysense\cite{Guo2023SkySenseAM} all the aforementioned works consists of models whose architectural number of parameters is less than $300$-million and are trained on datasets whose corpus is no where near the internet scale volumes encountered when training LLMs. Scaling the architectural sizes of FMs has been shown to bring forth many desirable benefits including task generalization and label efficiency. For these reasons, we contrast our work by introducing a systematic study profiling billion scale models pretrained on globally diversified high resolution satellite imagery.
\section{Methods}

All \textsc{OReole-MR/-HR} models are ``vanilla'' ViTs paired with MAE pretraining. We opt for such simpler configurations to focus on assessing model scaling effects. Future work will explore leveraging unique geospatial and temporal data characteristics, such as GSD-aware positional encodings \cite{reed2023scalemae}, contrastive learning using multiple collections over same location, and other architectural strategies such as RVSA \cite{wang2022rvsa}. 

\noindent\textbf{Model Architecture Variants} Table \ref{table:models} summarizes the different ViT variants explored in this work. \textit{Width} corresponds to embedding size, \textit{depth} corresponds to number of encoding layers, while \textit{heads} denote the number of heads per self-attention layer. While ViT-B configurations are paired with input patches $16\times16$ pixels large, all other larger ViT variants are paired with $14\times14$ pixels patches as per \cite{dosovitskiy2020vit} and related works. 
The study in \cite{dosovitskiy2020vit} discusses how scaling all aspects (i.e., depth, width, MLP-width, and patch-size) by similar amounts is most effective, while \cite{zhai2021scalingvit} performs extensive simulations to define ViT variants of different sizes. We follow these empirical guides and \cite{dehghani2023scaling} (ViT-22B) to increase the number of encoder layers and heads by gradually scaling the embedding size from $768$ to $2816$. The decoder pretraining architecture follows the original MAE work \cite{he2022masked}. 

\begin{table}[h]
\centering
\caption{ViT variants for \textsc{OReole} models} 
\begin{tabular}{ lccccc }
 \toprule
 \textbf{Model} & \textbf{Width}  & \textbf{Depth} & \textbf{MLP} & \textbf{Heads} & \textbf{\makecell{Parameters}} \\
 \midrule
 ViT-B &   768  & 12  & 3072 & 12  & 87M \\
 ViT-H &   1280  & 32  & 5120 & 16  & 635M \\
 ViT-G (1B) &   1536  & 32  & 6144 & 16  & 914M \\
 ViT-e (3B) &   2816  & 32  & 11264 & 32  & 3067M \\
 \bottomrule
\end{tabular}
\label{table:models}
\end{table}

\noindent\textbf{Scalability strategies} As part of this research, in \cite{tsaris2024pretraining} we discuss studies using the Frontier supercomputer on model sharding and data parallelism. Main observations include: i) FSDP-enabled implementations provide significant throughput gains when compared to PyTorch's DDP; ii) communication overheads rather than file I/O become the main bottleneck for such MAE-type of workloads. 
We refer the interested reader to \cite{tsaris2024pretraining} for further discussions on model sharding strategies and weak scaling studies. In the sections that follow, we adopt such a FSDP-enabled implementation for model pretraining. 

\noindent\textbf{Pretraining configuration}
Table \ref{tab:pretrain} summarizes the pretraining datasets and duration for \textsc{OReole} models and related works, which highly vary in pretraining durations (up to $1600M+$ iterations). We pretrain \textsc{OReole} models for $\approx200M$ iterations, as a trade-off between long enough pretraining to enable generalizable models while avoiding indiscriminate usage of computing resources. Details on code-bases used for pretraining are available in the Appendix.

\subsection{Pretraining datasets} 
Even though SSL removes the needs for data annotation, the acquisition of pretraining samples ought to be carefully designed. In the context of geospatial data, key characteristics to be considered during dataset curation include geographic diversity, richness, and scalability, to foment the learning of generic representations.



Examples of high-resolution RS datasets attempting to capture these characteristics include the MillionAID and the functional map of the world (fMoW). The fMoW dataset \cite{christie_fmow_2017} used by \cite{cong2022satmae, reed2023scalemae} contains 500k+ optical patches collected from multiple sensors. In contrast, the MillionAID \cite{long2021millionaid} used by \cite{wang2022rvsa, cha2023billion} contains $1M+$ RS scenes from a variety sensors and ground sample distances (GSD=0.5-150$m/px$). RingMo instead \cite{sun2022ringmo} exploits a not publicly-available dataset containing $2M+$ images for $419M$ iterations. 

Given the compute intense demands for pretraining FMs and the lack of standardized datasets, early studies have sought to pretrain on a single benchmark dataset\cite{reed2023scalemae, ayush2021geography, cha2023billion, wang2022rvsa}, while other studies merge multiple data sources to increase the volume and diversity of samples\cite{sun2022ringmo}. A growing body of work\cite{sumbul2021bigearthnet, Manas2021ICCV, Bastani2022SatlasPretrainAL, jakubik2023foundation, Guo2023SkySenseAM} is curating multimodal datasets sampled from large archives of remote sensing imagery including national agricultural imagery program (NAIP), Sentinel-1/-2, Harmonized Landsat Sentinel-2 (HLS), or WorldView -2/-3. We introduce two additional datasets in this paper, including the novel TIU.


\begin{table}[h]
    \centering
    \setlength{\tabcolsep}{3.5pt}
    \caption{Pretraining datasets and configurations.}
    \label{tab:pretrain}
    \small
\setlength{\tabcolsep}{3pt} 
\begin{tabular}{lccccccc}
    \toprule
    \textbf{Dataset} & \textbf{Bands} & \textbf{GSD} & \textbf{Volume} & \multicolumn{3}{c}{\textbf{Iterations}} \\
    \cmidrule(lr){5-7}
    & & \textbf{[m/px]} & \textbf{[$\sim$TB]}& \textbf{RVSA} & \textbf{BFM} & \textbf{OReole} \\
    \midrule
    \multirow{1}{*}{MillionAID} & RGB & 0.5-153 & 0.14 & 1601.4M & 400.3M & 200.2M \\
    \midrule
    \multirow{1}{*}{ORB} & BGR+NIR & $\approx$0.47 & 0.70 & - & - & 202.1M \\
    \multirow{1}{*}{TIU} & BGR+NIR & 0.3-0.8 & 1.30 & - & - & 202.1M \\
    \bottomrule
\end{tabular}    
\end{table}
\begin{figure}[h]
    \centering
    \includegraphics[width=.9\linewidth]{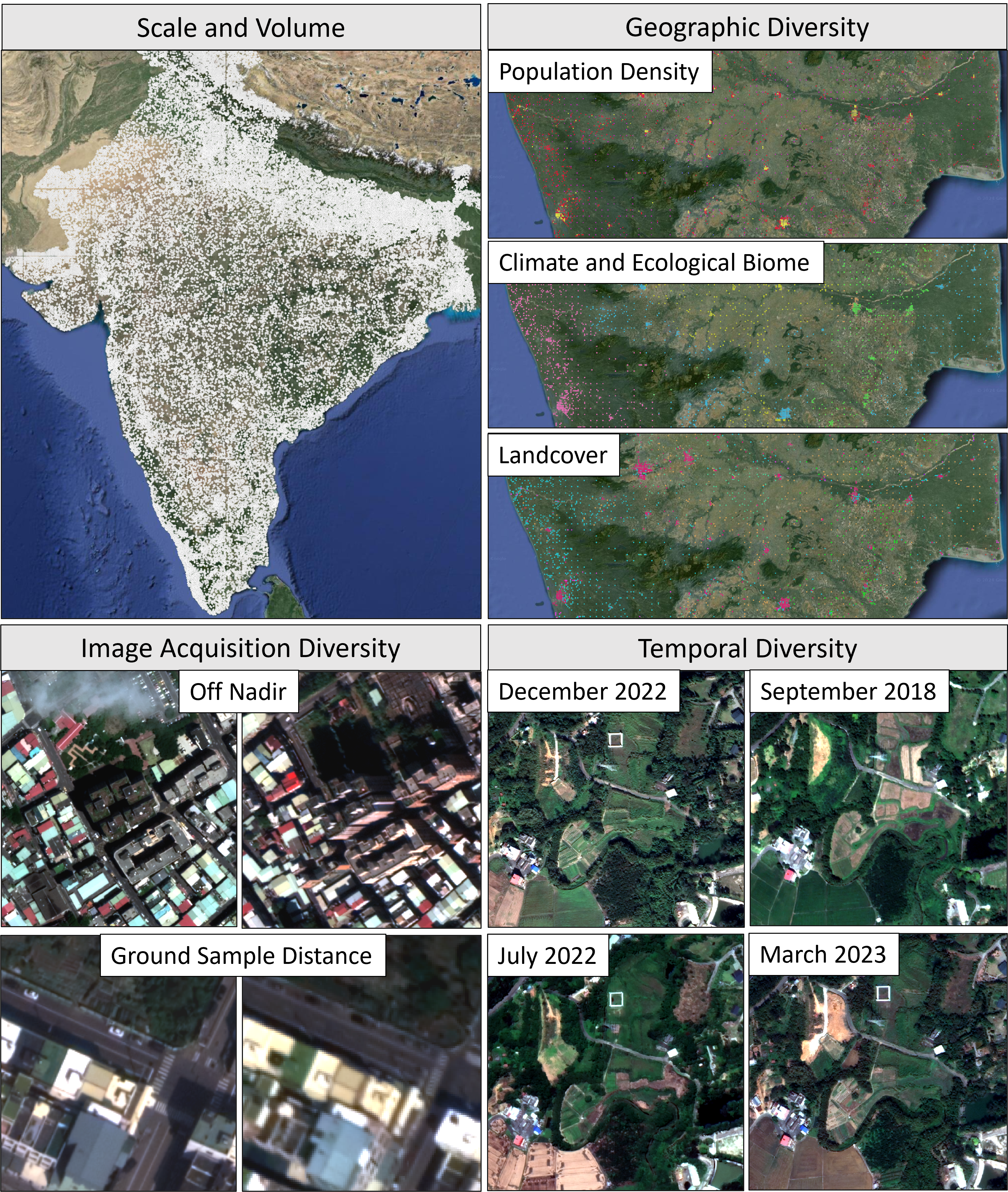}
    \caption{Illustration of characteristics of the \textit{TIU} dataset.}
    \label{fig:TIU-india-samples}
\end{figure}

\textbf{ORBITaL-Net.} ORBITaL-Net \cite{orbital-net} (ORB for shortness) is a semantic segmentation dataset for building footprint extraction (BFE) from high-resolution BGR+NIR imagery. The dataset includes $\approx126k$ samples collected across diverse geographic regions including North America, Asia, Africa, South America, and the Middle East.  Samples in this dataset consist of four channels (blue, green, red, near-infrared), were collected by the GeoEye-1, QuickBird-2, WorldView-2, and WorldView-3 sensors, have average ground sample distance of $0.47m/px$, and include binary masks to support the finetuning of FMs on a downstream building segmentation task.


\textbf{TIU.} TIU is an unlabeled dataset we constructed using the sampling strategy from \cite{arndt2024igarss}. It promotes diversity in:
i) \textit{geography} by referencing existing geographic information including population density, land cover, climate zones, and ecological biomes during sampling; ii) \textit{temporality} by including multiple views of the same location across different meteorological seasons (e.g. spring, summer, fall, winter) and years; iii) \textit{image acquisition} by including views from different sensors and viewing geometries. TIU consists of $\approx170k$ BGR+NIR images from GeoEye-1, WorldView-2/3 sensors, $1024\times1024px$ large at $GSD=$0.3-0.8m/px, sampled from $\approx80k$  locations across Taiwan, India, and Ukraine, with $\approx60\%$ of the locations with non-zero 24-hour population estimates. It contains up to four temporal views per location, each from a different meteorological season between years 2015 and 2023. Figure~\ref{fig:TIU-india-samples} illustrates characteristics of the dataset.

\label{sec:methods}

\section{\textsc{OReole-MR} Experiments}
\label{sec:experiments_mr}




We perform fine-tuning (FT) and linear probing (LP) experiments for comparison with existing baselines. Following the community trend, we consider image classification, object detection, and semantic segmentation downstream tasks. There is, however, a wide variability on the datasets used for performance benchmarking. We summarize in Table \ref{tab:eval_sets} the datasets commonly used to benchmark models pretrained on medium-to-high resolution imagery. Guided by this review, we selected the following datasets to compose the evaluation scenarios for \textsc{OReole-MR} variants: \textit{i) image classification:} UCM \cite{Yang2010UCM}, AID \cite{xia2017aid}, and NWPU\cite{cheng2017nwpu}; \textit{ii) object detection:} DIOR \cite{Li_2020} and DIOR-R \cite{Cheng_2022}; \textit{iii) semantic segmentation:} Potsdam and LoveDA \cite{wang2021loveda}. Details on the composition of these datasets and finetuning configurations are available in the Appendix. Results are summarized in Table \ref{tab:combined_results}, with main takeaways discussed below.
\begin{table}[h]
    \centering
    \caption{Summary of datasets used in related works for evaluation of downstream capabilities of pretrained models. In \textbf{bold}: the datasets selected for evaluation of \textsc{OReole-MR}.}
    \label{tab:eval_sets}
    \setlength{\tabcolsep}{1pt}
    \begin{tabular}{lccc}
    \cline{2-4}
         & \multicolumn{3}{c}{\textbf{Evaluation datasets}}\\
         \cline{2-4}
         &  \textit{\makecell{Image \\ Classification}} & \textit{\makecell{Object \\ Detection}}  & \textit{\makecell{Semantic \\ Segmentation}}\\
         \midrule
         RVSA&  \makecell{\textbf{UCM}, \textbf{AID},\\ \textbf{NWPU}}&  \makecell{\textbf{DIOR-R},\\ DOTA-V1.0}& \makecell{\textbf{Potsdam}, \\iSAID, \textbf{LoveDA}}\\[0.2cm]
         RingMo& \makecell{\textbf{UCM}, \textbf{AID},\\ \textbf{NWPU}}&  \makecell{\textbf{DIOR},\\ FAIR-1M}& \makecell{\textbf{Potsdam},\\ iSAID}\\[0.2cm]
         MTP&  \makecell{EuroSAT,\\ \textbf{NWPU}}&  \makecell{ \textbf{DIOR/-R}, FAIR1M,\\ DOTA V1.0/V2.0}& \makecell{SpaceNetv1, \\ \textbf{LoveDA}}\\[0.2cm]
 BFM& -& \makecell{\textbf{DIOR-R},\\ DOTA-V2.0} & \makecell{\textbf{Potsdam},\\ \textbf{LoveDA}}\\
 \midrule
 \textsc{Quetzal-MR} & \makecell{\textbf{UCM}, \textbf{AID},\\ \textbf{NWPU}} & \textbf{DIOR},\textbf{DIOR-R} & \makecell{\textbf{Potsdam},\\ \textbf{LoveDA}} \\
 \bottomrule
    \end{tabular}
\end{table}

\begin{table}[h]
\centering
\setlength{\tabcolsep}{3pt} 
\begin{threeparttable}
    \caption{Results across: image classification (top); object detection and semantic segmentation (bottom).}    
  \label{tab:combined_results}
  \setlength{\tabcolsep}{3.5pt} 
    \small\begin{tabular}{llcccccc}
    \toprule
     & \textbf{Model} & \textbf{\makecell{Pretrain \\ epochs}} & \textbf{\textit{\makecell{NWPU\\\small(TR=$10\%$)}}} & \multicolumn{2}{c}{\textbf{\textit{\makecell{UCM\\\small(TR=$50\%$)}}}} & \multicolumn{2}{c}{\textbf{\textit{\makecell{AID\\\small(TR=$20\%$)}}}} \\
    \midrule
        && & & \textbf{LP} & \textbf{FT} & \textbf{LP} & \textbf{FT} \\
    \multirow{3}{*}{\vspace{10pt}\rotatebox{90}{\textbf{\makecell{ViT-B \\ \cite{wang2022rvsa}}}}} & LR=0.1 & 1600 & 60.19 & 40.15 & -- & 60.17 & -- \\
    & LR=10  & 1600 & 87.20 & 98.58 & -- & 93.84 & -- \\
    \midrule
    \multirow{4}{*}{\rotatebox{90}{\textbf{\makecell{\textsc{OReole}\\\textsc{-MR}}}}} & ViT-B          & 200  & 86.19 & 97.35 & 99.05 & 93.70 & 94.73 \\
    & ViT-H          & 200  & 89.73 & 98.39 & 99.24 & 94.76 & 96.15 \\
    & ViT-G (1B)     & 200  & 79.54 & 94.22 & 99.24 & 89.38 & 96.71 \\
    & ViT-e (3B)     & 200  & 89.27 & 98.30 & -- & 95.53 & -- \\
    \bottomrule \vspace{0.05cm}
  \end{tabular}
\end{threeparttable}
\begin{threeparttable}
  \setlength{\tabcolsep}{3pt} 
  \small\begin{tabular}{llcccc}
    \toprule
     &  & \multicolumn{2}{c}{\textbf{Obj. detection}} & \multicolumn{2}{c}{\textbf{Segmentation}} \\
    \cmidrule(lr){3-4} \cmidrule(lr){5-6}
    \textbf{Method} & \textbf{Backbone} & \textbf{\small DIOR} & \textbf{\small DIOR-R} & \textbf{\small Potsdam} & \textbf{\small LoveDA} \\
    \textbf{} & \textbf{} & \multicolumn{2}{c}{\textbf{\small [mAP$_{50}$]}} & \textbf{\small [mF1 $\%$]} & \textbf{\small [mIoU $\%$]} \\
    \midrule
    GASSL & \small ResNet-50 & 67.40  & 65.65 & 91.27 & 48.76 \\
    MAE \small \cite{wang2022rvsa} & \small ViT-B & - & 66.65 & 90.32 & 51.09 \\
    RVSA \small \cite{wang2024mtp} & \small ViT-B+RVSA & 75.80  & 70.85 & 90.77 & 51.95 \\
    MTP \small \cite{wang2024mtp} & \small ViT-B+RVSA & 79.40 & - & - & 52.39 \\
    RVSA \small \cite{wang2024mtp} & \small ViT-L+RVSA & 78.30 & - & - & 53.72 \\
    MTP \small \cite{wang2024mtp} & \small ViT-L+RVSA & 81.10 & - & - & 54.17 \\
    SatMAE & \small ViT-L & 70.89  & 65.66 & 90.63 & - \\
    GFM \small 
    & \small Swin-B & 72.84 & 67.67 & 91.85 & - \\
    ScaleMAE & \small Swin-B & 73.81 & 66.47 & 91.54 & - \\
    RingMo(\small ISSP) & \small Swin-B & 73.50 & - & 91.27 & - \\
    RingMo \small \cite{sun2022ringmo} & \small Swin-B & 75.90 & - & 91.74 & - \\
    BFM \small \cite{cha2023billion} & \small ViT-B & - & 68.52 & - & - \\
    BFM \small \cite{cha2023billion} & \small ViT-L (605M) & - & 72.11 & 91.75 & 52.38 \\
    BFM \small \cite{cha2023billion} & \small ViT-H (1.36B) & - & 73.15 & 92.02 & 53.20 \\
    BFM \cite{cha2023billion} & \small ViT-G (2.42B) & - & 73.62 & 92.12 & 54.40 \\
    \midrule
    SkySense &\small  Swin-L & 76.74 & - & 92.86 & - \\
    SkySense & \small Swin-H & 78.73 & - & 93.99 & - \\
    \midrule
    \textsc{MAE\tnote{$\diamondsuit$}} & \small ViT-B & - & - & 91.82 & 52.97 \\
    \textsc{OReole-MR} & \small ViT-B & 75.50 & 68.58 & 92.01 & 52.44 \\
    \textsc{OReole-MR} & \small ViT-H & N/A \tnote{2} & N/A \tnote{2} & 92.18 & 53.83 \\
    \textsc{OReole-MR} & \small ViT-G (1B) & 77.40 & 71.31 & 92.20 & 54.00 \\
  \bottomrule
  \end{tabular}
  \begin{tablenotes}[flushleft]\footnotesize
    \item [1] Since GASSL, SatMAE, ScaleMAE, and GFM did not provide evaluation on DIOR(-R) and Potsdam in their original papers, we rely on results provided in \cite{wang2022rvsa,wang2024mtp,Guo2023SkySenseAM}.     
    \item[2] Results are not available due to convergence issues. 
  \end{tablenotes}  
\end{threeparttable}
\end{table}

\vspace{10pt}\noindent\textbf{Hyperparameters used by related works are often suboptimal, and may vary with model sizes and downstream tasks.} Similar to \cite{wang2022rvsa}, we opt for linear probing for image classification experiments, since these datasets have nearly saturated for FT. Critically, experiments using ViT-B pretrained checkpoints from \cite{wang2022rvsa} show that for the configurations reported in \cite{wang2022rvsa}, increasing the base learning rate (LR) from $0.1$ up to $10$ highly increases LP accuracy. 

Similarly, for object detection increasing the base LR of \textsc{OReole-MR} by $3\times$ improved ViT-G (1B) performance by nearly $+2.0\%$ in DIOR-R. Still, neither the configurations adopted for ViT-B nor ViT-G (1B) avoided loss divergence for ViT-H and ViT-e(3B) models. We conjecture two reasons: i) the object detection head is more complex, including RPN and RoI heads each with different regression losses (in contrast, e.g., to cross-entropy based optimization of UperNets for semantic segmentation); ii) related studies \cite{li2022exploring} highlight the need for different learning decay policies for different model sizes.

For semantic segmentation, \cite{wang2022rvsa, cha2023billion} employ $160k$ iterations (\textit{BS=8}) for both datasets, while \cite{sun2022ringmo} and \cite{Guo2023SkySenseAM} use $80k$. These setups equate to $1.28M$ and $0.64M$ iterations at \textit{BS=1}. For higher efficiency, we conducted experiments using distributed computing (effective \textit{BS=256}) for only $0.42M$ (Potsdam) and $0.346M$ (LoveDA) iterations at $BS=1$. Our ViT-B surpasses \cite{wang2022rvsa} by $>+1\%$ in both datasets, despite both pretrained and finetuned for $2x$ fewer iterations. Moreover, the MAE$^\diamondsuit$ resulting from finetuning RVSA ViT-B checkpoints \cite{wang2022rvsa} with our hyperparameters significantly surpasses results in \cite{wang2022rvsa}.

These observations indicate that configurations reported in the literature have been often suboptimal, and underscore an open challenge on how to establish standardized benchmarking practices.

\vspace{10pt}\noindent\textbf{Model scaling is beneficial, but with diminished gains without data scaling.}
For LP image classification with \textsc{OReole-MR} variants ($LR=10$), scaling from ViT-B(87M) to ViT-H (635M) yields gains, but they are diminished when scaling further into ViT-e (3B), with improvements only observed for AID. For object detection experiments, despite using a vanilla oriented-faster RCNN, for both DIOR and DIOR-R our ViT-B performs comparably to BFM, whose detection heads employ advanced ROI transformers. In general, the scaling benefits in model sizes (ViT-B to ViT-G) is clear for both datasets, and comparisons with BFE suggest the usage of ROI transformer provides further benefits when scaling models. 

For semantic segmentation, benefits of model scaling are particularly evident for LoveDA, but gains are diminished when scaling from ViT-H (635M) to ViT-G (1B). Considering discussions in the computer vision literature \cite{xie2023data} and \textsc{OReole-HR} results discussed in the next section, we conjecture 1B+ models require further data scaling. Contrasting datasets, the MillionAID used for \textsc{OReole-MR} is $10\times$ smaller than the TIU dataset that successfully enabled \textsc{OReole-HR}'s ViT-H. Pretraining duration is another potential factor: while \textsc{OReole-MR} ViT-B performs on par with BFM's, it is pretrained for $0.5\times$ iterations, and larger models have been shown to benefit more from data scaling when pretraining is longer \cite{xie2023data}. \textsc{OReole-MR}'s ViT-H (635M) variant also performs comparably to the BFM's 1B+ models \cite{cha2023billion}, representing a stronger baseline to improve upon.

\vspace{10pt}\noindent\textbf{MAE-learned features may have limited linear separability}
Surprisingly, ViT-G (1B) underperforms smaller models across all LP experiments. Since the same model however outperforms smaller configurations in other tasks, we conjectured the features learned through MAE pretraining have limited linear separability. To confirm, we ran FT image classification experiments that show instead saturation for UCM, and diminished yet positive gains for AID. This is consistent with reports in the literature of MAE often underperforming, e.g., CL in LP, since pixel-reconstruction pretraining foments focus on high-frequency details, rather than semantics.  

\vspace{10pt}\noindent\textbf{Many works lack reproducibility details for downstream evaluation}
Experiments using smaller portions of labeled data (available in supplementary material) highlight a lack of reproducibility details on related works. They often report maintaining the same FT configuration when varying data percentages, which is unclear when training schedules are reported in terms of epochs. We opted for maintaining the same number of training iterations for all data budgets, since the goal is to assess model's sample efficiency, not sensitivity to duration. Our results show model scaling benefits for as low as $>10\%$ training images of DIOR-R are used, and $1\%$ for segmentation experiments. Notably, our ViT-B at lower sample budgets  significantly surpass numbers in \cite{cha2023billion}, which we conjecture is due to our FT maintaining original training duration.

\section{\textsc{OReole-HR} experiments}
We leverage the ORB dataset for \textsc{OReole-HR} experiments. A separate random subset of $4,836$ tiles is used for validation, and two FT configurations were considered: \textit{$TR=100\%$}, with $55.71k$ labeled training tiles and FT for $100$ epochs ($5.57M$ iterations); \textit{$TR=10\%$}, using $5,571$ ($10\%$) randomly sampled tiles for FT ($100$ epochs = $557.1k$ iterations). Results are summarized in Table \ref{tab:bfe_semseg} and discussed below.

\begin{table}[h]
\small
  \caption{\textsc{OReole-HR} results on ORBiTal-Net dataset.}
  \setlength{\tabcolsep}{6pt}
  \label{tab:bfe_semseg}
  \begin{threeparttable}
    \begin{tabular}{llcccc}
    \toprule
    \textbf{Method} & \textbf{Pretrain} & \multicolumn{4}{c}{\bfseries BFE - mF1[\%]} \\ \cline{3-6} 
    & \textbf{dataset}& \multicolumn{2}{c}{\bfseries $TR=100\%$} & \multicolumn{2}{c}{\bfseries $TR=10\%$} \\ \cline{3-6}
    & & ViT-B & ViT-H & ViT-B & ViT-H \\
    \midrule
    No pretrain & - & 91.80 & - & 74.68 & - \\
    \midrule
    MAE & ORB & 91.79 & 91.31 & 89.57 & 88.59 \\ 
    \midrule
    MAE & TIU & 92.05 & 92.54 & 89.85 & 91.13 \\ 
    \midrule
    \textsc{O-MR}\tnote{1}+MAE & TIU & 92.15 & - & 90.17 & - \\
    \midrule
    MAE & TIU+ORB & 91.94 & 92.84 & 90.07 & 91.70 \\ 
    \midrule
    MAE ($224px^2$)& TIU+ORB & - & - & - & 90.38 \\ 
  \bottomrule
\end{tabular}

\begin{tablenotes}[flushleft]\footnotesize
    \item[1] initialized with MillionAID pretrained \textsc{OReole-MR} weights inflated to 4-bands.
\end{tablenotes}
\end{threeparttable}
\end{table}

\noindent\textbf{Pretraining on ORB dataset vs no pretrain (ViT-B)} While MAE pretraining on the same ORB dataset used for finetuning does not improve F1 in the $TR=100\%$ configuration, results in the $TR=10\%$ regime does reveal a significant benefit of pretraining in improving the time-to-solution (i.e. consumes less computing resources).

\vspace{10pt}\noindent\textbf{Model scaling requires data scaling, and larger models benefit more from data scaling} ViT-H models pretrained on the ORB dataset are worse than ViT-B for both TR configurations. In contrast, pretraining with the larger TIU dataset enables ViT-H model to significantly outperform ViT-B in both TRs configurations. Compared to ViT-B counterparts ($-0.11/+0.22\%$), the higher capacity ViT-H also benefits more from pretraining on the larger TIU+ORB dataset as compared to TIU only, with F1 for $TR=10\%$ similar to the ones obtained by ViT-B in $TR=100\%$ configuration.

\vspace{10pt}\noindent\textbf{Initialization using inflated \textsc{OReole-MR} weights.} Aiming speed-ups in training convergence and improved feature extraction, prior to MAE pretraining our O-MR+MAE configuration initializes a ViT-B with RGB \textsc{OReole-MR} pretrained weights, inflated with extra 4th-band randomly initialized weights. Results corroborate that the model's intrinsic exposure to MillionAID's images provide indirect data scaling benefits, with F1 rates boosted for both TR settings in comparison to TIU-only pretraining.  


\vspace{10pt}\noindent\textbf{Pretraining using larger image sizes} While pretraining a ViT-H using $224^2px$ instead of $512^2px$ patches reduces  GPU usage in $>4\times$, our experiments have shown a $-1.32\%$ F1 drop for $TR=10\%$ finetuning. This indicates that pretraining with larger patches can significantly benefit fine-grained tasks if computing can be afforded.

\section{Conclusion}

We introduced \textsc{OReole}, a family of FMs for RS image interpretation. Experiments with \textsc{OReole-MR} highlight model scaling benefits across downstream tasks, but with a pattern of diminished gains. Paired with takeaways from \textsc{OReole-HR} experiments, it highlights the importance of pairing model and data scaling, confirming for RS data a behavior observed in the broader computer vision \cite{xie2023data}.

Furthermore, we share a rare discussion on challenges related to downstream task optimization and setting up evaluation protocols. Many works lack reproducibility details such as the number of training iterations as dataset sizes vary. Moreover, we show how hyperparameters co-opted from related works can be largely suboptimal, and how parameters are sensitive to model size and tasks. This is a challenge for benchmarking, as exhaustive hyperparameter tuning is computationally expensive and contradictory to the intended benefits of FMs. We argue in favor of benchmarks with fixed-budget hyperparameter search, as suggested in  \cite{lacoste2023geobench}. 





Building on these conclusions and our work \cite{tsaris2024pretraining}, we aim to scale models beyond 3B parameters with expanded pretraining datasets, conduct few-shot evaluations to explore potential emergent abilities and implement geospatial-aware methods and integrate multiple data modalities to improve using FMs for Earth understanding.


\bibliographystyle{ACM-Reference-Format}
\bibliography{acmart}


\begin{thebibliography}{59}


\ifx \showCODEN    \undefined \def \showCODEN     #1{\unskip}     \fi
\ifx \showDOI      \undefined \def \showDOI       #1{#1}\fi
\ifx \showISBNx    \undefined \def \showISBNx     #1{\unskip}     \fi
\ifx \showISBNxiii \undefined \def \showISBNxiii  #1{\unskip}     \fi
\ifx \showISSN     \undefined \def \showISSN      #1{\unskip}     \fi
\ifx \showLCCN     \undefined \def \showLCCN      #1{\unskip}     \fi
\ifx \shownote     \undefined \def \shownote      #1{#1}          \fi
\ifx \showarticletitle \undefined \def \showarticletitle #1{#1}   \fi
\ifx \showURL      \undefined \def \showURL       {\relax}        \fi
\providecommand\bibfield[2]{#2}
\providecommand\bibinfo[2]{#2}
\providecommand\natexlab[1]{#1}
\providecommand\showeprint[2][]{arXiv:#2}

\bibitem[Fro({[n.\,d.]})]%
        {FrontierWebsite}
 \bibinfo{year}{[n.\,d.]}\natexlab{}.
\newblock \bibinfo{title}{{The Frontier supercomputer}}.
\newblock \bibinfo{howpublished}{\url{https://www.olcf.ornl.gov/frontier/}}.
\newblock


\bibitem[Arndt et~al\mbox{.}(2024)]%
        {arndt2024igarss}
\bibfield{author}{\bibinfo{person}{Jacob Arndt}, \bibinfo{person}{Philipe
  Dias}, \bibinfo{person}{Abhishek Potnis}, {and} \bibinfo{person}{Dalton
  Lunga}.} \bibinfo{year}{2024}\natexlab{}.
\newblock \showarticletitle{Towards Diverse and Representative Global
  Pretraining Datasets for Remote Sensing Foundation Models}. In
  \bibinfo{booktitle}{\emph{IEEE IGARSS}}. \bibinfo{pages}{2723--2728}.
\newblock


\bibitem[Arndt and Lunga(2021)]%
        {arndt2021urbanstructural}
\bibfield{author}{\bibinfo{person}{Jacob Arndt} {and} \bibinfo{person}{Dalton
  Lunga}.} \bibinfo{year}{2021}\natexlab{}.
\newblock \showarticletitle{Large-Scale Classification of Urban Structural
  Units From Remote Sensing Imagery}.
\newblock \bibinfo{journal}{\emph{IEEE Journal of Selected Topics in Applied
  Earth Observations and Remote Sensing}}  \bibinfo{volume}{14}
  (\bibinfo{year}{2021}), \bibinfo{pages}{2634--2648}.
\newblock
\urldef\tempurl%
\url{https://doi.org/10.1109/JSTARS.2021.3052961}
\showDOI{\tempurl}


\bibitem[Ayush et~al\mbox{.}(2021)]%
        {ayush2021geography}
\bibfield{author}{\bibinfo{person}{Kumar Ayush}, \bibinfo{person}{Burak
  Uzkent}, \bibinfo{person}{Chenlin Meng}, \bibinfo{person}{Kumar Tanmay},
  \bibinfo{person}{Marshall Burke}, \bibinfo{person}{David Lobell}, {and}
  \bibinfo{person}{Stefano Ermon}.} \bibinfo{year}{2021}\natexlab{}.
\newblock \showarticletitle{Geography-aware self-supervised learning}. In
  \bibinfo{booktitle}{\emph{IEEE/CVF ICCV}}. \bibinfo{pages}{10181--10190}.
\newblock


\bibitem[Bastani et~al\mbox{.}(2022)]%
        {Bastani2022SatlasPretrainAL}
\bibfield{author}{\bibinfo{person}{Favyen Bastani}, \bibinfo{person}{Piper
  Wolters}, \bibinfo{person}{Ritwik Gupta}, \bibinfo{person}{Joe Ferdinando},
  {and} \bibinfo{person}{Aniruddha Kembhavi}.} \bibinfo{year}{2022}\natexlab{}.
\newblock \showarticletitle{SatlasPretrain: A Large-Scale Dataset for Remote
  Sensing Image Understanding}.
\newblock \bibinfo{journal}{\emph{IEEE/CVF ICCV}} (\bibinfo{year}{2022}),
  \bibinfo{pages}{16726--16736}.
\newblock
\urldef\tempurl%
\url{https://api.semanticscholar.org/CorpusID:258947021}
\showURL{%
\tempurl}


\bibitem[Bhaduri et~al\mbox{.}(2007)]%
        {bhaduri2007landscan}
\bibfield{author}{\bibinfo{person}{Budhendra Bhaduri}, \bibinfo{person}{Edward
  Bright}, \bibinfo{person}{Phillip Coleman}, {and} \bibinfo{person}{Marie~L
  Urban}.} \bibinfo{year}{2007}\natexlab{}.
\newblock \showarticletitle{LandScan USA: a high-resolution geospatial and
  temporal modeling approach for population distribution and dynamics}.
\newblock \bibinfo{journal}{\emph{GeoJournal}}  \bibinfo{volume}{69}
  (\bibinfo{year}{2007}), \bibinfo{pages}{103--117}.
\newblock


\bibitem[Brown et~al\mbox{.}(2020)]%
        {brown2020gpt3}
\bibfield{author}{\bibinfo{person}{Tom Brown}, \bibinfo{person}{Benjamin Mann},
  \bibinfo{person}{Nick Ryder}, \bibinfo{person}{Melanie Subbiah},
  \bibinfo{person}{Jared~D Kaplan}, \bibinfo{person}{Prafulla Dhariwal},
  \bibinfo{person}{Arvind Neelakantan}, \bibinfo{person}{Pranav Shyam},
  \bibinfo{person}{Girish Sastry}, \bibinfo{person}{Amanda Askell},
  {et~al\mbox{.}}} \bibinfo{year}{2020}\natexlab{}.
\newblock \showarticletitle{Language models are few-shot learners}.
\newblock \bibinfo{journal}{\emph{Advances in neural information processing
  systems}}  \bibinfo{volume}{33} (\bibinfo{year}{2020}),
  \bibinfo{pages}{1877--1901}.
\newblock


\bibitem[Cha et~al\mbox{.}(2023)]%
        {cha2023billion}
\bibfield{author}{\bibinfo{person}{Keumgang Cha}, \bibinfo{person}{Junghoon
  Seo}, {and} \bibinfo{person}{Taekyung Lee}.} \bibinfo{year}{2023}\natexlab{}.
\newblock \showarticletitle{A billion-scale foundation model for remote sensing
  images}.
\newblock \bibinfo{journal}{\emph{arXiv:2304.05215}} (\bibinfo{year}{2023}).
\newblock


\bibitem[Chen et~al\mbox{.}(2019)]%
        {mmdetection}
\bibfield{author}{\bibinfo{person}{Kai Chen}, \bibinfo{person}{Jiaqi Wang},
  \bibinfo{person}{Jiangmiao Pang}, \bibinfo{person}{Yuhang Cao},
  \bibinfo{person}{Yu Xiong}, \bibinfo{person}{Xiaoxiao Li},
  \bibinfo{person}{Shuyang Sun}, \bibinfo{person}{Wansen Feng},
  \bibinfo{person}{Ziwei Liu}, \bibinfo{person}{Jiarui Xu},
  \bibinfo{person}{Zheng Zhang}, \bibinfo{person}{Dazhi Cheng},
  \bibinfo{person}{Chenchen Zhu}, \bibinfo{person}{Tianheng Cheng},
  \bibinfo{person}{Qijie Zhao}, \bibinfo{person}{Buyu Li}, \bibinfo{person}{Xin
  Lu}, \bibinfo{person}{Rui Zhu}, \bibinfo{person}{Yue Wu},
  \bibinfo{person}{Jifeng Dai}, \bibinfo{person}{Jingdong Wang},
  \bibinfo{person}{Jianping Shi}, \bibinfo{person}{Wanli Ouyang},
  \bibinfo{person}{Chen~Change Loy}, {and} \bibinfo{person}{Dahua Lin}.}
  \bibinfo{year}{2019}\natexlab{}.
\newblock \showarticletitle{{MMDetection}: Open MMLab Detection Toolbox and
  Benchmark}.
\newblock \bibinfo{journal}{\emph{arXiv:1906.07155}} (\bibinfo{year}{2019}).
\newblock


\bibitem[Chen et~al\mbox{.}(2020)]%
        {chen2020simclr}
\bibfield{author}{\bibinfo{person}{Ting Chen}, \bibinfo{person}{Simon
  Kornblith}, \bibinfo{person}{Mohammad Norouzi}, {and}
  \bibinfo{person}{Geoffrey Hinton}.} \bibinfo{year}{2020}\natexlab{}.
\newblock \showarticletitle{A simple framework for contrastive learning of
  visual representations}. In \bibinfo{booktitle}{\emph{International
  conference on machine learning}}. PMLR, \bibinfo{pages}{1597--1607}.
\newblock


\bibitem[Cheng et~al\mbox{.}(2017)]%
        {cheng2017nwpu}
\bibfield{author}{\bibinfo{person}{Gong Cheng}, \bibinfo{person}{Junwei Han},
  {and} \bibinfo{person}{Xiaoqiang Lu}.} \bibinfo{year}{2017}\natexlab{}.
\newblock \showarticletitle{Remote Sensing Image Scene Classification:
  Benchmark and State of the Art}.
\newblock \bibinfo{journal}{\emph{IEEE}} \bibinfo{volume}{105},
  \bibinfo{number}{10} (\bibinfo{year}{2017}), \bibinfo{pages}{1865--1883}.
\newblock


\bibitem[Cheng et~al\mbox{.}(2022)]%
        {Cheng_2022}
\bibfield{author}{\bibinfo{person}{Gong Cheng}, \bibinfo{person}{Jiabao Wang},
  \bibinfo{person}{Ke Li}, \bibinfo{person}{Xingxing Xie},
  \bibinfo{person}{Chunbo Lang}, \bibinfo{person}{Yanqing Yao}, {and}
  \bibinfo{person}{Junwei Han}.} \bibinfo{year}{2022}\natexlab{}.
\newblock \showarticletitle{Anchor-Free Oriented Proposal Generator for Object
  Detection}.
\newblock \bibinfo{journal}{\emph{IEEE Trans. on Geoscience and Remote
  Sensing}}  \bibinfo{volume}{60} (\bibinfo{year}{2022}),
  \bibinfo{pages}{1–11}.
\newblock
\showISSN{1558-0644}


\bibitem[Chowdhery et~al\mbox{.}(2023)]%
        {chowdhery2023palm}
\bibfield{author}{\bibinfo{person}{Aakanksha Chowdhery},
  \bibinfo{person}{Sharan Narang}, \bibinfo{person}{Jacob Devlin},
  \bibinfo{person}{Maarten Bosma}, \bibinfo{person}{Gaurav Mishra},
  \bibinfo{person}{Adam Roberts}, \bibinfo{person}{Paul Barham},
  \bibinfo{person}{Hyung~Won Chung}, \bibinfo{person}{Charles Sutton},
  \bibinfo{person}{Sebastian Gehrmann}, {et~al\mbox{.}}}
  \bibinfo{year}{2023}\natexlab{}.
\newblock \showarticletitle{Palm: Scaling language modeling with pathways}.
\newblock \bibinfo{journal}{\emph{Journal of Machine Learning Research}}
  \bibinfo{volume}{24}, \bibinfo{number}{240} (\bibinfo{year}{2023}),
  \bibinfo{pages}{1--113}.
\newblock


\bibitem[Christie et~al\mbox{.}(2018)]%
        {christie_fmow_2017}
\bibfield{author}{\bibinfo{person}{Gordon Christie}, \bibinfo{person}{Neil
  Fendley}, \bibinfo{person}{James Wilson}, {and} \bibinfo{person}{Ryan
  Mukherjee}.} \bibinfo{year}{2018}\natexlab{}.
\newblock \showarticletitle{Functional map of the world}. In
  \bibinfo{booktitle}{\emph{IEEE CVPR}}. \bibinfo{pages}{6172--6180}.
\newblock


\bibitem[Cong et~al\mbox{.}(2022)]%
        {cong2022satmae}
\bibfield{author}{\bibinfo{person}{Yezhen Cong}, \bibinfo{person}{Samar
  Khanna}, \bibinfo{person}{Chenlin Meng}, \bibinfo{person}{Patrick Liu},
  \bibinfo{person}{Erik Rozi}, {et~al\mbox{.}}}
  \bibinfo{year}{2022}\natexlab{}.
\newblock \showarticletitle{Satmae: Pre-training transformers for temporal and
  multi-spectral satellite imagery}.
\newblock \bibinfo{journal}{\emph{Advances in Neural Information Processing
  Systems}}  \bibinfo{volume}{35} (\bibinfo{year}{2022}),
  \bibinfo{pages}{197--211}.
\newblock


\bibitem[Contributors(2023)]%
        {2023mmpretrain}
\bibfield{author}{\bibinfo{person}{MMPreTrain Contributors}.}
  \bibinfo{year}{2023}\natexlab{}.
\newblock \bibinfo{title}{OpenMMLab's Pre-training Toolbox and Benchmark}.
\newblock
  \bibinfo{howpublished}{\url{https://github.com/open-mmlab/mmpretrain}}.
\newblock


\bibitem[Dehghani et~al\mbox{.}(2023)]%
        {dehghani2023scaling}
\bibfield{author}{\bibinfo{person}{Mostafa Dehghani}, \bibinfo{person}{Josip
  Djolonga}, \bibinfo{person}{Basil Mustafa}, \bibinfo{person}{Piotr
  Padlewski}, \bibinfo{person}{Jonathan Heek}, \bibinfo{person}{Justin Gilmer},
  \bibinfo{person}{Andreas~Peter Steiner}, \bibinfo{person}{Mathilde Caron},
  \bibinfo{person}{Robert Geirhos}, \bibinfo{person}{Ibrahim Alabdulmohsin},
  {et~al\mbox{.}}} \bibinfo{year}{2023}\natexlab{}.
\newblock \showarticletitle{Scaling vision transformers to 22 billion
  parameters}. In \bibinfo{booktitle}{\emph{International Conference on Machine
  Learning}}. PMLR, \bibinfo{pages}{7480--7512}.
\newblock


\bibitem[Devlin et~al\mbox{.}(2018)]%
        {devlin2018bert}
\bibfield{author}{\bibinfo{person}{Jacob Devlin}, \bibinfo{person}{Ming-Wei
  Chang}, \bibinfo{person}{Kenton Lee}, {and} \bibinfo{person}{Kristina
  Toutanova}.} \bibinfo{year}{2018}\natexlab{}.
\newblock \showarticletitle{Bert: Pre-training of deep bidirectional
  transformers for language understanding}.
\newblock \bibinfo{journal}{\emph{arXiv:1810.04805}} (\bibinfo{year}{2018}).
\newblock


\bibitem[Dias et~al\mbox{.}(2024)]%
        {dias2024bdicond}
\bibfield{author}{\bibinfo{person}{Philipe Dias}, \bibinfo{person}{Jacob
  Arndt}, \bibinfo{person}{Marie Urban}, {and} \bibinfo{person}{Dalton Lunga}.}
  \bibinfo{year}{2024}\natexlab{}.
\newblock \showarticletitle{Conditional Experts for Improved Building Damage
  Assessment Across Satellite Imagery View Angles}. In
  \bibinfo{booktitle}{\emph{IGARSS 2024 - 2024 IEEE International Geoscience
  and Remote Sensing Symposium}}. \bibinfo{pages}{1741--1745}.
\newblock
\urldef\tempurl%
\url{https://doi.org/10.1109/IGARSS53475.2024.10640461}
\showDOI{\tempurl}


\bibitem[Dosovitskiy et~al\mbox{.}(2020)]%
        {dosovitskiy2020vit}
\bibfield{author}{\bibinfo{person}{Alexey Dosovitskiy}, \bibinfo{person}{Lucas
  Beyer}, \bibinfo{person}{Alexander Kolesnikov}, \bibinfo{person}{Dirk
  Weissenborn}, \bibinfo{person}{Xiaohua Zhai}, \bibinfo{person}{Thomas
  Unterthiner}, \bibinfo{person}{Mostafa Dehghani}, \bibinfo{person}{Matthias
  Minderer}, \bibinfo{person}{Georg Heigold}, \bibinfo{person}{Sylvain Gelly},
  {et~al\mbox{.}}} \bibinfo{year}{2020}\natexlab{}.
\newblock \showarticletitle{An Image is Worth 16x16 Words: Transformers for
  Image Recognition at Scale}. In \bibinfo{booktitle}{\emph{International
  Conference on Learning Representations}}.
\newblock


\bibitem[{GDAL/OGR contributors}(2024)]%
        {gdal}
\bibfield{author}{\bibinfo{person}{{GDAL/OGR contributors}}.}
  \bibinfo{year}{2024}\natexlab{}.
\newblock \bibinfo{booktitle}{\emph{{GDAL/OGR} Geospatial Data Abstraction
  software Library}}.
\newblock Open Source Geospatial Foundation.
\newblock
\urldef\tempurl%
\url{https://doi.org/10.5281/zenodo.5884351}
\showDOI{\tempurl}


\bibitem[Girshick(2015)]%
        {7410526}
\bibfield{author}{\bibinfo{person}{Ross Girshick}.}
  \bibinfo{year}{2015}\natexlab{}.
\newblock \showarticletitle{Fast R-CNN}. In \bibinfo{booktitle}{\emph{2015 IEEE
  International Conference on Computer Vision (ICCV)}}.
  \bibinfo{pages}{1440--1448}.
\newblock
\urldef\tempurl%
\url{https://doi.org/10.1109/ICCV.2015.169}
\showDOI{\tempurl}


\bibitem[Guo et~al\mbox{.}(2023)]%
        {Guo2023SkySenseAM}
\bibfield{author}{\bibinfo{person}{Xin Guo}, \bibinfo{person}{Jiangwei Lao},
  \bibinfo{person}{Bo Dang}, \bibinfo{person}{Yingying Zhang},
  \bibinfo{person}{Lei Yu}, {et~al\mbox{.}}} \bibinfo{year}{2023}\natexlab{}.
\newblock \showarticletitle{SkySense: A Multi-Modal Remote Sensing Foundation
  Model Towards Universal Interpretation for Earth Observation Imagery}.
\newblock \bibinfo{journal}{\emph{ArXiv}}  \bibinfo{volume}{abs/2312.10115}
  (\bibinfo{year}{2023}).
\newblock


\bibitem[He et~al\mbox{.}(2022)]%
        {he2022masked}
\bibfield{author}{\bibinfo{person}{Kaiming He}, \bibinfo{person}{Xinlei Chen},
  \bibinfo{person}{Saining Xie}, \bibinfo{person}{Yanghao Li},
  \bibinfo{person}{Piotr Doll{\'a}r}, {and} \bibinfo{person}{Ross Girshick}.}
  \bibinfo{year}{2022}\natexlab{}.
\newblock \showarticletitle{Masked autoencoders are scalable vision learners}.
  In \bibinfo{booktitle}{\emph{IEEE/CVF CVPR}}. \bibinfo{pages}{16000--16009}.
\newblock


\bibitem[Hong et~al\mbox{.}(2023)]%
        {Hong2023SpectralGPTSR}
\bibfield{author}{\bibinfo{person}{Danfeng Hong}, \bibinfo{person}{Bing Zhang},
  \bibinfo{person}{Xuyang Li}, \bibinfo{person}{Yuxuan Li},
  \bibinfo{person}{Chenyu Li}, \bibinfo{person}{Jing Yao},
  \bibinfo{person}{Naoto Yokoya}, \bibinfo{person}{Hao Li},
  \bibinfo{person}{Pedram Ghamisi}, \bibinfo{person}{Xiuping Jia},
  \bibinfo{person}{Antonio~J. Plaza}, \bibinfo{person}{Paolo Gamba},
  \bibinfo{person}{J{\'o}n~Atli Benediktsson}, {and} \bibinfo{person}{Jocelyn
  Chanussot}.} \bibinfo{year}{2023}\natexlab{}.
\newblock \showarticletitle{SpectralGPT: Spectral Remote Sensing Foundation
  Model.}
\newblock \bibinfo{journal}{\emph{IEEE Trans. on pattern analysis and machine
  intelligence}}  \bibinfo{volume}{PP} (\bibinfo{year}{2023}).
\newblock
\urldef\tempurl%
\url{https://api.semanticscholar.org/CorpusID:267628000}
\showURL{%
\tempurl}


\bibitem[Irvin et~al\mbox{.}(2023)]%
        {irvin2023usat}
\bibfield{author}{\bibinfo{person}{Jeremy Irvin}, \bibinfo{person}{Lucas Tao},
  \bibinfo{person}{Joanne Zhou}, \bibinfo{person}{Yuntao Ma},
  \bibinfo{person}{Langston Nashold}, \bibinfo{person}{Benjamin Liu}, {and}
  \bibinfo{person}{Andrew~Y Ng}.} \bibinfo{year}{2023}\natexlab{}.
\newblock \showarticletitle{USat: A unified self-supervised encoder for
  multi-sensor satellite imagery}.
\newblock \bibinfo{journal}{\emph{arXiv preprint arXiv:2312.02199}}
  (\bibinfo{year}{2023}).
\newblock


\bibitem[Jakubik et~al\mbox{.}(2023)]%
        {jakubik2023foundation}
\bibfield{author}{\bibinfo{person}{Johannes Jakubik}, \bibinfo{person}{Sujit
  Roy}, \bibinfo{person}{C.~E. Phillips}, \bibinfo{person}{Paolo Fraccaro},
  \bibinfo{person}{Denys Godwin}, \bibinfo{person}{Bianca Zadrozny},
  \bibinfo{person}{Daniela Szwarcman}, \bibinfo{person}{Carlos Gomes},
  \bibinfo{person}{Gabby Nyirjesy}, \bibinfo{person}{Blair Edwards},
  \bibinfo{person}{Daiki Kimura}, \bibinfo{person}{Naomi Simumba},
  \bibinfo{person}{Linsong Chu}, \bibinfo{person}{S.~Karthik Mukkavilli},
  \bibinfo{person}{Devyani Lambhate}, \bibinfo{person}{Kamal Das},
  \bibinfo{person}{Ranjini Bangalore}, \bibinfo{person}{Dario Oliveira},
  \bibinfo{person}{Michal Muszynski}, \bibinfo{person}{Kumar Ankur},
  \bibinfo{person}{Muthukumaran Ramasubramanian}, \bibinfo{person}{Iksha
  Gurung}, \bibinfo{person}{Sam Khallaghi}, \bibinfo{person}{Hanxi},
  \bibinfo{person}{Li}, \bibinfo{person}{Michael Cecil},
  \bibinfo{person}{Maryam Ahmadi}, \bibinfo{person}{Fatemeh Kordi},
  \bibinfo{person}{Hamed Alemohammad}, \bibinfo{person}{Manil Maskey},
  \bibinfo{person}{Raghu Ganti}, \bibinfo{person}{Kommy Weldemariam}, {and}
  \bibinfo{person}{Rahul Ramachandran}.} \bibinfo{year}{2023}\natexlab{}.
\newblock \bibinfo{title}{Foundation Models for Generalist Geospatial
  Artificial Intelligence}.
\newblock
\newblock
\showeprint[arxiv]{2310.18660}~[cs.CV]


\bibitem[Lacoste et~al\mbox{.}(2024)]%
        {lacoste2023geobench}
\bibfield{author}{\bibinfo{person}{Alexandre Lacoste}, \bibinfo{person}{Nils
  Lehmann}, \bibinfo{person}{Pau Rodriguez}, \bibinfo{person}{Evan Sherwin},
  \bibinfo{person}{Hannah Kerner}, {et~al\mbox{.}}}
  \bibinfo{year}{2024}\natexlab{}.
\newblock \showarticletitle{Geo-bench: Toward foundation models for earth
  monitoring}.
\newblock \bibinfo{journal}{\emph{Advances in Neural Information Processing
  Systems}}  \bibinfo{volume}{36} (\bibinfo{year}{2024}).
\newblock


\bibitem[Laverdiere et~al\mbox{.}(2020)]%
        {laverdiere2020rapid}
\bibfield{author}{\bibinfo{person}{Melanie Laverdiere}, \bibinfo{person}{Lexie
  Yang}, \bibinfo{person}{Mark Tuttle}, {and} \bibinfo{person}{Chris Vaughan}.}
  \bibinfo{year}{2020}\natexlab{}.
\newblock \showarticletitle{Rapid Structure Detection in Support of Disaster
  Response: A Case Study of the 2018 Kilauea Volcano Eruption}. In
  \bibinfo{booktitle}{\emph{IGARSS 2020 - 2020 IEEE International Geoscience
  and Remote Sensing Symposium}}. \bibinfo{pages}{6826--6829}.
\newblock
\urldef\tempurl%
\url{https://doi.org/10.1109/IGARSS39084.2020.9324160}
\showDOI{\tempurl}


\bibitem[Li et~al\mbox{.}(2020)]%
        {Li_2020}
\bibfield{author}{\bibinfo{person}{Ke Li}, \bibinfo{person}{Gang Wan},
  \bibinfo{person}{Gong Cheng}, \bibinfo{person}{Liqiu Meng}, {and}
  \bibinfo{person}{Junwei Han}.} \bibinfo{year}{2020}\natexlab{}.
\newblock \showarticletitle{Object detection in optical remote sensing images:
  A survey and a new benchmark}.
\newblock \bibinfo{journal}{\emph{ISPRS Journal of Photogrammetry and Remote
  Sensing}}  \bibinfo{volume}{159} (\bibinfo{year}{2020}),
  \bibinfo{pages}{296–307}.
\newblock
\showISSN{0924-2716}


\bibitem[Li et~al\mbox{.}(2022)]%
        {li2022exploring}
\bibfield{author}{\bibinfo{person}{Yanghao Li}, \bibinfo{person}{Hanzi Mao},
  \bibinfo{person}{Ross Girshick}, {and} \bibinfo{person}{Kaiming He}.}
  \bibinfo{year}{2022}\natexlab{}.
\newblock \showarticletitle{Exploring plain vision transformer backbones for
  object detection}. In \bibinfo{booktitle}{\emph{ECCV}}.
  \bibinfo{pages}{280--296}.
\newblock


\bibitem[Long et~al\mbox{.}(2021)]%
        {long2021millionaid}
\bibfield{author}{\bibinfo{person}{Yang Long}, \bibinfo{person}{Gui-Song Xia},
  \bibinfo{person}{Shengyang Li}, \bibinfo{person}{Wen Yang},
  \bibinfo{person}{Michael~Ying Yang}, \bibinfo{person}{Xiao~Xiang Zhu},
  \bibinfo{person}{Liangpei Zhang}, {and} \bibinfo{person}{Deren Li}.}
  \bibinfo{year}{2021}\natexlab{}.
\newblock \showarticletitle{On creating benchmark dataset for aerial image
  interpretation: Reviews, guidances, and million-aid}.
\newblock \bibinfo{journal}{\emph{IEEE Journal of selected topics in applied
  earth observations and remote sensing}}  \bibinfo{volume}{14},
  \bibinfo{pages}{4205--4230}.
\newblock


\bibitem[Lunga et~al\mbox{.}(2021)]%
        {lunga2021resflow}
\bibfield{author}{\bibinfo{person}{Dalton Lunga}, \bibinfo{person}{Jacob
  Arndt}, \bibinfo{person}{Jonathan Gerrand}, {and} \bibinfo{person}{Robert
  Stewart}.} \bibinfo{year}{2021}\natexlab{}.
\newblock \showarticletitle{ReSFlow: A Remote Sensing Imagery Data-Flow for
  Improved Model Generalization}.
\newblock \bibinfo{journal}{\emph{IEEE Journal of Selected Topics in Applied
  Earth Observations and Remote Sensing}}  \bibinfo{volume}{14}
  (\bibinfo{year}{2021}), \bibinfo{pages}{10468--10483}.
\newblock
\urldef\tempurl%
\url{https://doi.org/10.1109/JSTARS.2021.3119001}
\showDOI{\tempurl}


\bibitem[Ma\~nas et~al\mbox{.}(2021)]%
        {Manas2021ICCV}
\bibfield{author}{\bibinfo{person}{Oscar Ma\~nas}, \bibinfo{person}{Alexandre
  Lacoste}, \bibinfo{person}{Xavier Gir\'o-i Nieto}, \bibinfo{person}{David
  Vazquez}, {and} \bibinfo{person}{Pau Rodr{\'\i}guez}.}
  \bibinfo{year}{2021}\natexlab{}.
\newblock \showarticletitle{Seasonal Contrast: Unsupervised Pre-Training From
  Uncurated Remote Sensing Data}. In \bibinfo{booktitle}{\emph{IEEE/CVF ICCV}}.
  \bibinfo{pages}{9414--9423}.
\newblock


\bibitem[Mendieta et~al\mbox{.}(2023)]%
        {mendieta2023towards}
\bibfield{author}{\bibinfo{person}{Mat{\'\i}as Mendieta},
  \bibinfo{person}{Boran Han}, \bibinfo{person}{Xingjian Shi},
  \bibinfo{person}{Yi Zhu}, {and} \bibinfo{person}{Chen Chen}.}
  \bibinfo{year}{2023}\natexlab{}.
\newblock \showarticletitle{Towards geospatial foundation models via continual
  pretraining}. In \bibinfo{booktitle}{\emph{IEEE/CVF ICCV}}.
  \bibinfo{pages}{16806--16816}.
\newblock


\bibitem[Radford et~al\mbox{.}({[n.\,d.]})]%
        {radford2018gpt}
\bibfield{author}{\bibinfo{person}{Alec Radford}, \bibinfo{person}{Karthik
  Narasimhan}, \bibinfo{person}{Tim Salimans}, \bibinfo{person}{Ilya
  Sutskever}, {et~al\mbox{.}}} \bibinfo{year}{[n.\,d.]}\natexlab{}.
\newblock \showarticletitle{Improving language understanding by generative
  pre-training}.
\newblock  (\bibinfo{year}{[n.\,d.]}).
\newblock


\bibitem[Reed et~al\mbox{.}(2023)]%
        {reed2023scalemae}
\bibfield{author}{\bibinfo{person}{Colorado~J Reed}, \bibinfo{person}{Ritwik
  Gupta}, \bibinfo{person}{Shufan Li}, \bibinfo{person}{Sarah Brockman},
  \bibinfo{person}{Christopher Funk}, {et~al\mbox{.}}}
  \bibinfo{year}{2023}\natexlab{}.
\newblock \showarticletitle{Scale-mae: A scale-aware masked autoencoder for
  multiscale geospatial representation learning}. In
  \bibinfo{booktitle}{\emph{IEEE/CVF ICCV}}. \bibinfo{pages}{4088--4099}.
\newblock


\bibitem[Sims et~al\mbox{.}(2023)]%
        {sims2023landscan}
\bibfield{author}{\bibinfo{person}{Kelly Sims}, \bibinfo{person}{Andrew Reith},
  \bibinfo{person}{Elizabeth Bright}, \bibinfo{person}{Jon Kaufman},
  \bibinfo{person}{Jordan Pyle}, \bibinfo{person}{Julie Epting},
  \bibinfo{person}{Juan Gonzales}, \bibinfo{person}{David Adams},
  \bibinfo{person}{Emily Powell}, \bibinfo{person}{Matthew Urban}, {and}
  \bibinfo{person}{Adam Rose}.} \bibinfo{year}{2023}\natexlab{}.
\newblock \bibinfo{title}{LandScan Global 2022}.
\newblock \bibinfo{howpublished}{\url{https://doi.org/10.48690/1529167}}.
\newblock


\bibitem[Sumbul et~al\mbox{.}(2021)]%
        {sumbul2021bigearthnet}
\bibfield{author}{\bibinfo{person}{Gencer Sumbul}, \bibinfo{person}{Arne
  De~Wall}, \bibinfo{person}{Tristan Kreuziger}, \bibinfo{person}{Filipe
  Marcelino}, \bibinfo{person}{Hugo Costa}, \bibinfo{person}{Pedro Benevides},
  \bibinfo{person}{Mario Caetano}, \bibinfo{person}{Beg{\"u}m Demir}, {and}
  \bibinfo{person}{Volker Markl}.} \bibinfo{year}{2021}\natexlab{}.
\newblock \showarticletitle{BigEarthNet-MM: A Large-Scale, Multimodal,
  Multilabel Benchmark Archive for Remote Sensing Image Classification and
  Retrieval [Software and Data Sets]}.
\newblock \bibinfo{journal}{\emph{IEEE Geoscience and Remote Sensing Magazine}}
  \bibinfo{volume}{9}, \bibinfo{number}{3} (\bibinfo{year}{2021}),
  \bibinfo{pages}{174--180}.
\newblock


\bibitem[Sun et~al\mbox{.}(2022)]%
        {sun2022ringmo}
\bibfield{author}{\bibinfo{person}{Xian Sun}, \bibinfo{person}{Peijin Wang},
  \bibinfo{person}{Wanxuan Lu}, \bibinfo{person}{Zicong Zhu},
  \bibinfo{person}{Xiaonan Lu}, \bibinfo{person}{Qibin He},
  \bibinfo{person}{Junxi Li}, \bibinfo{person}{Xuee Rong},
  \bibinfo{person}{Zhujun Yang}, {et~al\mbox{.}}}
  \bibinfo{year}{2022}\natexlab{}.
\newblock \showarticletitle{Ringmo: A remote sensing foundation model with
  masked image modeling}.
\newblock \bibinfo{journal}{\emph{IEEE Trans. Geosci. Remote Sens.,}}
  (\bibinfo{year}{2022}).
\newblock


\bibitem[Swan et~al\mbox{.}(2024)]%
        {orbital-net}
\bibfield{author}{\bibinfo{person}{Benjamin Swan}, \bibinfo{person}{Joe Pyle},
  \bibinfo{person}{Darrell Roddy}, \bibinfo{person}{Amy Rose},
  \bibinfo{person}{Lexie~H. Yang}, {and} \bibinfo{person}{Melanie Laverdiere}.}
  \bibinfo{year}{2024}\natexlab{}.
\newblock \showarticletitle{ORBITaL-Net Training Library for Building
  Extraction}.
\newblock \bibinfo{journal}{\emph{Figshare+ Dataset}} (\bibinfo{year}{2024}).
\newblock
\urldef\tempurl%
\url{https://doi.org/10.25452/figshare.plus.25282225.v1}
\showDOI{\tempurl}


\bibitem[Touvron et~al\mbox{.}(2023)]%
        {touvron2023llama}
\bibfield{author}{\bibinfo{person}{Hugo Touvron}, \bibinfo{person}{Thibaut
  Lavril}, \bibinfo{person}{Gautier Izacard}, \bibinfo{person}{Xavier
  Martinet}, \bibinfo{person}{Marie-Anne Lachaux},
  \bibinfo{person}{Timoth{\'e}e Lacroix}, \bibinfo{person}{Baptiste
  Rozi{\`e}re}, \bibinfo{person}{Naman Goyal}, \bibinfo{person}{Eric Hambro},
  \bibinfo{person}{Faisal Azhar}, {et~al\mbox{.}}}
  \bibinfo{year}{2023}\natexlab{}.
\newblock \showarticletitle{Llama: Open and efficient foundation language
  models}.
\newblock \bibinfo{journal}{\emph{arXiv:2302.13971}} (\bibinfo{year}{2023}).
\newblock


\bibitem[Tsaris et~al\mbox{.}(2024)]%
        {tsaris2024pretraining}
\bibfield{author}{\bibinfo{person}{Aristeidis Tsaris}, \bibinfo{person}{Philipe
  Dias}, \bibinfo{person}{Abhishek Potnis}, \bibinfo{person}{Junqi Yin},
  \bibinfo{person}{Feiyi Wang}, {and} \bibinfo{person}{Dalton Lunga}.}
  \bibinfo{year}{2024}\natexlab{}.
\newblock \showarticletitle{Pretraining Billion-Scale Geospatial Foundational
  Models on Frontier}. In \bibinfo{booktitle}{\emph{IEEE IPDPS Workshops}}.
  \bibinfo{pages}{1036--1046}.
\newblock


\bibitem[Vaswani et~al\mbox{.}(2017)]%
        {vaswani2017transformers}
\bibfield{author}{\bibinfo{person}{Ashish Vaswani}, \bibinfo{person}{Noam
  Shazeer}, \bibinfo{person}{Niki Parmar}, \bibinfo{person}{Jakob Uszkoreit},
  \bibinfo{person}{Llion Jones}, \bibinfo{person}{Aidan~N Gomez},
  \bibinfo{person}{{\L}ukasz Kaiser}, {and} \bibinfo{person}{Illia
  Polosukhin}.} \bibinfo{year}{2017}\natexlab{}.
\newblock \showarticletitle{Attention is all you need}.
\newblock \bibinfo{journal}{\emph{Advances in neural information processing
  systems}}  \bibinfo{volume}{30} (\bibinfo{year}{2017}).
\newblock


\bibitem[Wang et~al\mbox{.}(2022a)]%
        {wang2022empirical}
\bibfield{author}{\bibinfo{person}{Di Wang}, \bibinfo{person}{Jing Zhang},
  \bibinfo{person}{Bo Du}, \bibinfo{person}{Gui-Song Xia}, {and}
  \bibinfo{person}{Dacheng Tao}.} \bibinfo{year}{2022}\natexlab{a}.
\newblock \showarticletitle{An empirical study of remote sensing pretraining}.
\newblock \bibinfo{journal}{\emph{IEEE Trans. on Geoscience and Remote
  Sensing}} (\bibinfo{year}{2022}).
\newblock


\bibitem[Wang et~al\mbox{.}(2024)]%
        {wang2024mtp}
\bibfield{author}{\bibinfo{person}{Di Wang}, \bibinfo{person}{Jing Zhang},
  \bibinfo{person}{Minqiang Xu}, \bibinfo{person}{Lin Liu},
  \bibinfo{person}{Dongsheng Wang}, \bibinfo{person}{Erzhong Gao},
  \bibinfo{person}{Chengxi Han}, {et~al\mbox{.}}}
  \bibinfo{year}{2024}\natexlab{}.
\newblock \showarticletitle{MTP: Advancing Remote Sensing Foundation Model via
  Multi-Task Pretraining}.
\newblock \bibinfo{journal}{\emph{arXiv:2403.13430}} (\bibinfo{year}{2024}).
\newblock


\bibitem[Wang et~al\mbox{.}(2022b)]%
        {wang2022rvsa}
\bibfield{author}{\bibinfo{person}{Di Wang}, \bibinfo{person}{Qiming Zhang},
  \bibinfo{person}{Yufei Xu}, \bibinfo{person}{Jing Zhang}, \bibinfo{person}{Bo
  Du}, \bibinfo{person}{Dacheng Tao}, {and} \bibinfo{person}{Liangpei Zhang}.}
  \bibinfo{year}{2022}\natexlab{b}.
\newblock \showarticletitle{Advancing plain vision transformer toward remote
  sensing foundation model}.
\newblock \bibinfo{journal}{\emph{IEEE Trans. on Geoscience and Remote
  Sensing}}  \bibinfo{volume}{61} (\bibinfo{year}{2022}),
  \bibinfo{pages}{1--15}.
\newblock


\bibitem[Wang et~al\mbox{.}(2021)]%
        {wang2021loveda}
\bibfield{author}{\bibinfo{person}{Junjue Wang}, \bibinfo{person}{Zhuo Zheng},
  \bibinfo{person}{Ailong Ma}, \bibinfo{person}{Xiaoyan Lu}, {and}
  \bibinfo{person}{Yanfei Zhong}.} \bibinfo{year}{2021}\natexlab{}.
\newblock \showarticletitle{LoveDA: A Remote Sensing Land-Cover Dataset for
  Domain Adaptive Semantic Segmentation}. In \bibinfo{booktitle}{\emph{Neural
  Information Processing Systems Datasets and Benchmarks}}.
\newblock


\bibitem[Wei et~al\mbox{.}(2022)]%
        {wei2022emergent}
\bibfield{author}{\bibinfo{person}{Jason Wei}, \bibinfo{person}{Yi Tay},
  \bibinfo{person}{Rishi Bommasani}, \bibinfo{person}{Colin Raffel},
  \bibinfo{person}{Barret Zoph}, \bibinfo{person}{Sebastian Borgeaud},
  \bibinfo{person}{Dani Yogatama}, \bibinfo{person}{Maarten Bosma},
  \bibinfo{person}{Denny Zhou}, \bibinfo{person}{Donald Metzler},
  {et~al\mbox{.}}} \bibinfo{year}{2022}\natexlab{}.
\newblock \showarticletitle{Emergent abilities of large language models}.
\newblock \bibinfo{journal}{\emph{arXiv:2206.07682}} (\bibinfo{year}{2022}).
\newblock


\bibitem[Xia et~al\mbox{.}(2017)]%
        {xia2017aid}
\bibfield{author}{\bibinfo{person}{Gui-Song Xia}, \bibinfo{person}{Jingwen Hu},
  \bibinfo{person}{Fan Hu}, \bibinfo{person}{Baoguang Shi},
  \bibinfo{person}{Xiang Bai}, {et~al\mbox{.}}}
  \bibinfo{year}{2017}\natexlab{}.
\newblock \showarticletitle{AID: A Benchmark Data Set for Performance
  Evaluation of Aerial Scene Classification}.
\newblock \bibinfo{journal}{\emph{IEEE Trans. on Geoscience and Remote
  Sensing}} \bibinfo{volume}{55}, \bibinfo{number}{7},
  \bibinfo{pages}{3965--3981}.
\newblock


\bibitem[Xiao et~al\mbox{.}(2018)]%
        {xiao2018upernet}
\bibfield{author}{\bibinfo{person}{Tete Xiao}, \bibinfo{person}{Yingcheng Liu},
  \bibinfo{person}{Bolei Zhou}, \bibinfo{person}{Yuning Jiang}, {and}
  \bibinfo{person}{Jian Sun}.} \bibinfo{year}{2018}\natexlab{}.
\newblock \showarticletitle{Unified perceptual parsing for scene
  understanding}. In \bibinfo{booktitle}{\emph{European conference on computer
  vision (ECCV)}}. \bibinfo{pages}{418--434}.
\newblock


\bibitem[Xie et~al\mbox{.}(2023)]%
        {xie2023data}
\bibfield{author}{\bibinfo{person}{Zhenda Xie}, \bibinfo{person}{Zheng Zhang},
  \bibinfo{person}{Yue Cao}, \bibinfo{person}{Yutong Lin},
  \bibinfo{person}{Yixuan Wei}, {et~al\mbox{.}}}
  \bibinfo{year}{2023}\natexlab{}.
\newblock \showarticletitle{On data scaling in masked image modeling}. In
  \bibinfo{booktitle}{\emph{IEEE/CVF CVPR}}. \bibinfo{pages}{10365--10374}.
\newblock


\bibitem[Yang et~al\mbox{.}(2018)]%
        {yang2018building}
\bibfield{author}{\bibinfo{person}{Hsiuhan~Lexie Yang},
  \bibinfo{person}{Jiangye Yuan}, \bibinfo{person}{Dalton Lunga},
  \bibinfo{person}{Melanie Laverdiere}, \bibinfo{person}{Amy Rose}, {and}
  \bibinfo{person}{Budhendra Bhaduri}.} \bibinfo{year}{2018}\natexlab{}.
\newblock \showarticletitle{Building extraction at scale using convolutional
  neural network: Mapping of the united states}.
\newblock \bibinfo{journal}{\emph{IEEE Journal of Selected Topics in Applied
  Earth Observations and Remote Sensing}} \bibinfo{volume}{11},
  \bibinfo{number}{8} (\bibinfo{year}{2018}), \bibinfo{pages}{2600--2614}.
\newblock


\bibitem[{Yang} et~al\mbox{.}(2021)]%
        {yang2021agu}
\bibfield{author}{\bibinfo{person}{Lexie {Yang}}, \bibinfo{person}{Dalton
  {Lunga}}, \bibinfo{person}{Dawn {King}}, \bibinfo{person}{Jacob {Arndt}},
  \bibinfo{person}{Jordan {Bowman}}, {and} \bibinfo{person}{Robert {Stewart}}.}
  \bibinfo{year}{2021}\natexlab{}.
\newblock \showarticletitle{{Exploring Spatially Distributed Deep Learning
  Models for Global Gravitational Mapping}}. In \bibinfo{booktitle}{\emph{AGU
  Fall Meeting Abstracts}}, Vol.~\bibinfo{volume}{2021}. Article
  \bibinfo{articleno}{EP12C-05}, \bibinfo{numpages}{EP12C-05}~pages.
\newblock


\bibitem[Yang and Newsam(2010)]%
        {Yang2010UCM}
\bibfield{author}{\bibinfo{person}{Yi Yang} {and} \bibinfo{person}{Shawn
  Newsam}.} \bibinfo{year}{2010}\natexlab{}.
\newblock \showarticletitle{Bag-of-visual-words and spatial extensions for
  land-use classification}. In \bibinfo{booktitle}{\emph{ACM SIGSPATIAL
  Workshop on Advances in GIS}}.
\newblock


\bibitem[You et~al\mbox{.}(2017)]%
        {you2017large}
\bibfield{author}{\bibinfo{person}{Yang You}, \bibinfo{person}{Igor Gitman},
  {and} \bibinfo{person}{Boris Ginsburg}.} \bibinfo{year}{2017}\natexlab{}.
\newblock \bibinfo{title}{Large Batch Training of Convolutional Networks}.
\newblock
\newblock
\showeprint[arxiv]{1708.03888}~[cs.CV]


\bibitem[Zhai et~al\mbox{.}(2022)]%
        {zhai2021scalingvit}
\bibfield{author}{\bibinfo{person}{Xiaohua Zhai}, \bibinfo{person}{Alexander
  Kolesnikov}, \bibinfo{person}{Neil Houlsby}, {and} \bibinfo{person}{Lucas
  Beyer}.} \bibinfo{year}{2022}\natexlab{}.
\newblock \showarticletitle{Scaling vision transformers}. In
  \bibinfo{booktitle}{\emph{IEEE/CVF CVPR}}. \bibinfo{pages}{12104--12113}.
\newblock


\bibitem[Zhao et~al\mbox{.}(2023)]%
        {zhao2023pytorch}
\bibfield{author}{\bibinfo{person}{Yanli Zhao}, \bibinfo{person}{Andrew Gu},
  \bibinfo{person}{Rohan Varma}, \bibinfo{person}{Liang Luo},
  \bibinfo{person}{Chien-Chin Huang}, \bibinfo{person}{Min Xu},
  \bibinfo{person}{Less Wright}, \bibinfo{person}{Hamid Shojanazeri},
  \bibinfo{person}{Myle Ott}, \bibinfo{person}{Sam Shleifer},
  \bibinfo{person}{Alban Desmaison}, \bibinfo{person}{Can Balioglu},
  \bibinfo{person}{Pritam Damania}, \bibinfo{person}{Bernard Nguyen},
  \bibinfo{person}{Geeta Chauhan}, \bibinfo{person}{Yuchen Hao},
  \bibinfo{person}{Ajit Mathews}, {and} \bibinfo{person}{Shen Li}.}
  \bibinfo{year}{2023}\natexlab{}.
\newblock \bibinfo{title}{PyTorch FSDP: Experiences on Scaling Fully Sharded
  Data Parallel}.
\newblock
\newblock
\showeprint[arxiv]{2304.11277}~[cs.DC]


\bibitem[Zhou et~al\mbox{.}(2022)]%
        {zhou2022mmrotate}
\bibfield{author}{\bibinfo{person}{Yue Zhou}, \bibinfo{person}{Xue Yang},
  \bibinfo{person}{Gefan Zhang}, \bibinfo{person}{Jiabao Wang},
  \bibinfo{person}{Yanyi Liu}, \bibinfo{person}{Liping Hou},
  \bibinfo{person}{Xue Jiang}, \bibinfo{person}{Xingzhao Liu},
  \bibinfo{person}{Junchi Yan}, \bibinfo{person}{Chengqi Lyu},
  \bibinfo{person}{Wenwei Zhang}, {and} \bibinfo{person}{Kai Chen}.}
  \bibinfo{year}{2022}\natexlab{}.
\newblock \showarticletitle{MMRotate: A Rotated Object Detection Benchmark
  using PyTorch}. In \bibinfo{booktitle}{\emph{30th ACM International
  Conference on Multimedia}}.
\newblock


\end{thebibliography}
\newpage
\onecolumn

\section{Appendix}

\subsection{Pretraining implementation details}
Development of \textsc{OReole-MR} and \textsc{Oreole-HR} models took place concomitantly. \textsc{Oreole-HR} models were pretrained using the \textit{mmpretrain} codebase \cite{2023mmpretrain}, an open source pre-training toolbox based on PyTorch adapted using the gdal library \cite{gdal} for ingestion of the BGR+NIR images in TIF-file format composing our $TIU$ and $ORB$ datasets. We opted for the \textit{mmpretrain} for two main reasons. First, it contains implementations of a wide variety of model backbones and self-supervised learning strategies, offering flexibility for experimentation with other types of configurations in the long term. Moreover, it is part of the broader collection of repositories by OpenMMLab (e.g., \textit{mmsegmentation}, \textit{mmdetection} and \textit{mmrotate}), thus potentially facilitating integration with configurations for downstream tasks.

\textsc{OReole-HR} models were all trained with an effective batch size of $4096$ samples, distributed across GPUs composing nodes of the Frontier Supercomputer \cite{FrontierWebsite}. Each Frontier node contains $4\times$ AMD Instinct MI250X GPU accelerators, with each MI250X comprising of $2\times$ Graphics Compute Dies (GCDs). Since the system identifies each GCD independently, from the application perspective it can be considered that each node has a total of 8 GPUs, each with 64 GB of high-bandwidth memory. For simplicity we use the term GPU when referring to a GCD. ViT-H model variants were trained using $16$ Frontier nodes ($128$ GPUs), while ViT-B variants were trained with $8$ nodes ($64$ GPUs).

Meanwhile, \textsc{OReole-MR} models were pretrained using a codebase based on the original MAE work \cite{he2022masked}\footnote{https://github.com/facebookresearch/mae}. Specifically, we leverage the codebase and insights discussed in our recent work \cite{tsaris2024pretraining}, which augments the original MAE codebase to enable the support for PyTorch's native Fully Sharded Data Parallel (FSDP) strategy \cite{zhao2023pytorch}. Our studies detailed in \cite{tsaris2024pretraining} revealed a higher image throughput and hence time to solution for MAE pretraining when using FSDP NO\textunderscore{}SHARD strategies as compared to PyTorch's Distributed Data Parallel (DDP). We leveraged this knowledge for pretrained each of our \textsc{Oreole-MR} models using the FSDP NO\textunderscore{}SHARD strategy to distribute an effective batch size of $2048$ across 4 NVIDIA A100 (80GB) GPUs composing NVIDIA DGX clusters. 
\begin{table}[!htbp]
    \centering
    \caption{Implementation details adopted for pretraining of \textsc{OReole-MR} and \textsc{OReole-HR} model variants.}
    \label{tab:hyperparams_pre}    
    \begin{threeparttable}
    \begin{tabular}{ccc}
        \toprule
         \textbf{Details} & \textbf{\textsc{OReole-MR}} & \textbf{\textsc{OReole-HR}} \\
         \midrule
         Optimizer & \makecell{AdamW \\$\beta=(0.9, 0.95)$} & \makecell{AdamW \\$\beta=(0.9, 0.999)$} \\
         LR\tnote{1} &  1.2e-3 & 2.4e-3 \\ 
         LR scheduler & CosineAnnealingLR & CosineAnnealingLR\\
         Effective BS&  2048 & 4096 \\
         Warmup&  40 epochs & 40 epochs\\
         Image size&  224x224& 512x512 \\
         Image bands & RGB & BGR+NIR \\ 
         Weight Decay&  0.05& 0.05\\      
         Augmentation & \makecell{RandomResizedCrop \\ (crop ratio 0.2-1.0)\\RandomHorizontalFlip} & \makecell{RandomResizedCrop \\ (crop ratio 0.2-1.0)\\RandomHorizontalFlip} \\            
    \bottomrule
    \end{tabular}    
    \begin{tablenotes}\footnotesize
        \item[1] \textit{base\textunderscore{}lr} is set to $1.5e-4$ and scaled as $lr=base\_lr\times4096/256$ following \cite{he2022masked}; 
    \end{tablenotes}
    \end{threeparttable}
\end{table}
\subsection{Image classification experiments details}

\noindent\textbf{Datasets} \textit{UC Merced (UCM)} is a image classification dataset containing $2100$ remote sensing images from the USGS National Map, at a resolution of $1ft$. Images are $256\times256px$ large, distributed evenly across 21 categories. The \textit{AID} dataset was compiled from multisensor data available in Google Earth, with spatial resolutions $0.5 - 8m/px$ for images $600\times600px$ large. It contains $10k$ images distributed across 30 categories, with class distributions ranging from 220 to 400 images each. Finally, the \textit{NWPU-RESISC45} dataset contains $31.5k$ images evenly distributed across 45 categories and also collected from Google Earth for regions distributed across the world. Image are $256\times256px$ large, with spatial resolutions ranging from $0.2$ to $30m/px$. We adopt the following splits for each dataset: $UCM (TR=50\%)$ with 1050/1050 train/test images; AID ($TR=20\%$) with $2000/8000$ images;  NWPU ($TR=10\%$) with $3150/28350$ images.

\noindent\textbf{Implementation details} The MLP heads of pretrained models are replaced by a linear classifier for supervised training while keeping the weights for the rest of the model frozen. The setup in \cite{wang2022rvsa} reports experiments for $200$ epochs of tuning with batch-size $256$ for all three datasets while adapting usage of the LARS optimizer\cite{you2017large} with a base learning rate of 0.1 and no weight decay. As discussed in Section 4, we empirically observe that while keeping the same batch size, increasing the base learning up to $10$ in this setup yields severe improvements in accuracy levels reached after linear probing. Thus, for linear probing evaluation with \textsc{Quetzal-MR} model variants we adopt LARS with base learning rate $10$, tuning the learnable heads for $100$ epochs with a warmup period of $10$ epochs. 

\subsection{Object detection experiments details}
\noindent\textbf{Datasets}
\textit{DIOR} \cite{Li_2020} contains 23,463 $800\times800px$ RGB images sourced from Google Earth, with $GSD=0.5-30m/px$. There are 20 object classes and a total of 192,472 object instances. DIOR is divided into training, validation, and testing sets with $5,862$, $5,863$, and $11,738 images$, respectively. As in \cite{wang2024mtp}, we use the training and validation sets in a combined $11,725$-image "trainval" set for fine-tuning, and we validate using the testing set.  The images, object classes, object instances, and data splits are identical to DIOR.

\noindent\textbf{Implementation details} We augment our pretrained ViT backbones weights into VitDet \cite{li2022exploring}, a ViT-based object detection architecture that demonstrates promising performance with minimal adaptation of backbones required. Faster-RCNN \cite{7410526} was used as the detection head for the horizontal object detection task, and its variant Oriented Faster-RCNN was used for the oriented object detection task. The VitDet implementation was based on \textit{mmdetection} \cite{mmdetection} for DIOR data, and the same implementation was further adapted for DIOR-R data using the \textit{mmrotate} \cite{zhou2022mmrotate} library. We report evaluation based on Mean Average Precision (mAP) calculated at Intersection over Union (IoU) threshold 50\%.
    
As in \cite{li2022exploring}, we adopt a patch size of $16\times16px$ for all ViT backbones, interpolating the originally $14\times14px$ patch embeddings pretrained for the ViT-H and ViT-1B backbones. With the pretrained ViT weights, we finetuned on DIOR and DIOR-R for 12 epochs, the same as the setting in \cite{wang2024mtp}. Similar to \cite{wang2022rvsa}, the base learning rate was set as $1e-4$ for ViT-B with a batch size of 4. Experimentation with $[1, 3, 5, 10]e-4$ and same batch size for ViT-G has shown better convergence with $3e-4$, while for ViT-B the original $1e-4$ revealed to be better. The remaining configurations follow \cite{wang2024mtp} and are listed in the Table \ref{tab:hyperparams_ft}. Convergence issues are faced with ViT-H and ViT-e(3B) configurations, as curves in Figure \ref{fig:objdet-conv} show. As discussed in Section \ref{sec:experiments_mr}, we conjecture the complexity of the object detection head and the need for different learning decay policies for different model sizes \cite{li2022exploring} are potential root causes. 

\begin{table}[h]
    \centering
    \caption{Summary of hyperparameters and additional configuration details adopted for linear probing and finetuning experiments.}
    \setlength{\tabcolsep}{2.5pt}
    \label{tab:hyperparams_ft}
    \begin{tabular}{lcccccc}
    \toprule
 & \multicolumn{3}{c}{\textbf{Image classification}} & \multicolumn{1}{c}{\textbf{Object detection}} & \multicolumn{2}{c}{\textbf{Semantic segmentation}}\\
        \toprule
         \textbf{Details} & \textit{UCM} & \textit{AID} & \textit{NWPU} & \textit{DIOR/-R} & \textit{Potsdam} & \textit{LoveDA} \\
         \midrule
 Head/decoder & Linear Probing & Linear Probing & Linear Probing & Faster/Oriented Faster RCNN & UperNet+FCN & UperNet+FCN \\         
         Optimizer & LARS & LARS & LARS & AdamW & AdamW & AdamW \\
         Input size & 256 & 600 & 256 & 800 & 512 & 1024 \\
 Training samples & 1,050 & 2,000 & 3,150 & 11,725 & 3,456 & 4,191 \\
 Test samples & 1,050 & 8,000 & 28,350 & 11,738 & 2,016 & 1,796 \\
 \makecell[l]{Effective \\ batch size} & 256 & 256 & 256 & 4 & 256 & 256 \\
         Base LR & 10 & 10 & 10 & \makecell{ViT-B: 1e-4\\ ViT-H,1B,3B: 3e-4} & 5e-4 & 5e-4 \\[0.2cm]
 Warmup & 10 epochs & 10 epochs & 10 epochs & 500 iter & 50 epochs & 50 epochs \\[0.2cm]
         LR scheduler & \multicolumn{3}{c}{CosineAnnealingLR} & \makecell{StepLR: \\ $\gamma=0.1$, \\ decay at epochs $[8, 11]$} & CosineAnnealingLR & CosineAnnealingLR \\[0.2cm]
         Weight Decay & 0 & 0 & 0 & 0.05 & 0.05 & 0.05 \\
 Duration & 100 epochs & 100 epochs & 100 epochs & 12 epochs & 100 epochs & 100 epochs \\
 \makecell[l]{LR layer wise \\ decay rate} & 0.9 & 0.9 & 0.9 & 0.9 & 0.9 & 0.9\\[0.2cm]
 Drop path rate & -& -& -& 0.1 & 0. & 0 \\         
 Augmentation & \multicolumn{3}{c}{\makecell{RandomResizedCrop (224) \\ RandomHorizontalFlip}} & RandomFlip & \multicolumn{2}{c}{\makecell{RandomResize (0.5-2.0) \\ RandomCrop \\ RandomFlip \\ PhotoMetricDistortion}} \\
 Loss function & CE & CE & CE & CE & \multicolumn{2}{c}{\makecell{CE \\ FCN weight: 0.4}}\\
    \bottomrule
    \end{tabular}
\end{table}

\subsection{Semantic segmentation experiments details}\label{sec:semseg}

\noindent\textbf{Datasets.} The LoveDA \cite{wang2021loveda} comprises of 5987 high resolution imagery ($1024\times1024px$, $GSD=0.3m/px$) from Google Earth, with per-pixel annotations of 7 land cover classes. Images are split into train/val/test sets comprising 2522/1669/1796 samples respectively. The evaluation metrics have been computed on the test set using the evaluation server. The ISPRS \textit{Potsdam} dataset consists of 38 tiles of $GSD=0.5m/px$ orthoimagery, annotated with 6 land cover classes. Like related works, we tile images into $512\times512px$ patches, with train/test sets comprising of 3456/2016 patches.

\noindent\textbf{Implementation details.}
Following common practice across related works \cite{wang2022empirical, sun2022ringmo, Guo2023SkySenseAM}, we adopt the UperNet scheme \cite{xiao2018upernet} to compose segmentation frameworks for all our considered pretrained backbones. All experiments are implemented using the \textit{mmsegmentation} codebase. Following \cite{wang2022empirical, wang2022rvsa}, we configured the UperNet architectures with a Feature2Pyramid neck operating with $[4, 2, 1, 0.5]$ rescaling factors. For ViT-B configurations, features are collected from blocks $[3, 5, 7, 11]$ (index starting at 0) of the $12-$layer deep backbone. Since the ViT-H and ViT-g(1B) variants are $32-$ layers deep, we instead collect features from blocks $[11, 17, 23, 31]$. Moreover, a FCN with $256$ channels and loss weight of $0.4$ is used as auxiliary head. While we adopt such architectural configuration for consistency with the literature, experiments with the ViT-e(3B) remain as future work since its pairing UperNet + auxiliary FCN head does not fit in a single GPU for our setups, requiring further development using model sharding.

Configurations reported in \cite{wang2022rvsa} and \cite{cha2023billion} employ $160k$ iterations at batch size $8$, base LR of $6e-5$ and weight decay of $0.01$ for Potsdam and LoveDA, while \cite{sun2022ringmo} and \cite{Guo2023SkySenseAM} adopt $80k$ iterations for the same set of hyperparameters on Potsdam. This corresponds to finetuning durations of $1.28M$ and $0.64M$ iterations at batch-size $1$. To efficiently evaluate different model sizes across several configurations, we conduct these experiments under a distributed computing setup where such a small batch size is sub-optimal. Following experimentation with base learning rates for ViT-B on Potsdam, for all our semantic segmentation experiments we adopt an effective batch-size of $256$, with a base learning rate of $5e-4$ paired with CosineAnnealingLR scheduled with $50$ epochs of warmup and finetuning duration of $100$ epochs for each dataset. This corresponds to $345.6k$ iterations (at batch size $1$) for Potsdam and $419.1k$ iterations for LoveDA, which is equivalent to only $27\%$ of the iterations reported by \cite{wang2022rvsa, cha2023billion} for Potsdam and $65.5\%$ of the duration reported in \cite{wang2022rvsa} for LoveDA.

\subsection{\textsc{OReole-HR} experiments details}
Similar to experiments reported in section \ref{sec:semseg}, UperNet is paired to pretrained \textsc{OReole-HR} variants for semantic segmentation. Feature pyramids and rescaling factors are the same as reported in section \ref{sec:experiments_mr}, with the auxiliary FCN head however removed to reduce computing workload. We adopt distance-labels as per \cite{yang2018building} to formulate the loss function, and experiments are performed with a effective batch size of $2048$, AdamW with betas ($0.9$, $0.999$), base learning rate of $1e-3$, and a cosine annealing LR schedule with $10$ epochs of warmup. Image size of $512\times512$ is adopted for Vit-B configurations and $504\times504$ for larger configurations trained with patches $14\times14px$ large, to bypass mmsegmentation errors related to non-integer number of patches.


\subsection{\textsc{OReole-MR} - additional results}
\begin{figure*}[h]
    \centering
     \begin{subfigure}[b]{0.32\linewidth}
     \centering
        \includegraphics[clip, trim=0 0.45cm 0 0.1cm,width=\linewidth, height=2.5cm]{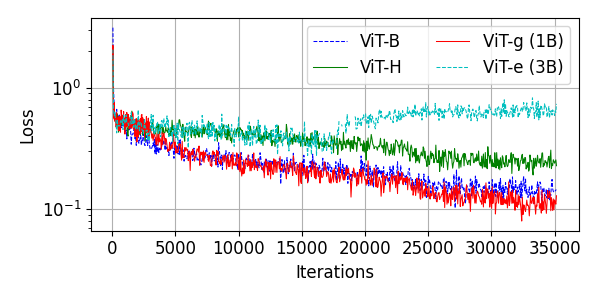}
         \caption{Training loss curves - (DIOR-R)}
         \label{fig:objdet-conv}
     \end{subfigure}      
     \begin{subfigure}[b]{0.32\linewidth}
     \centering
        \includegraphics[clip, trim=0 0.45cm 0 0.1cm,width=\linewidth, height=2.5cm]{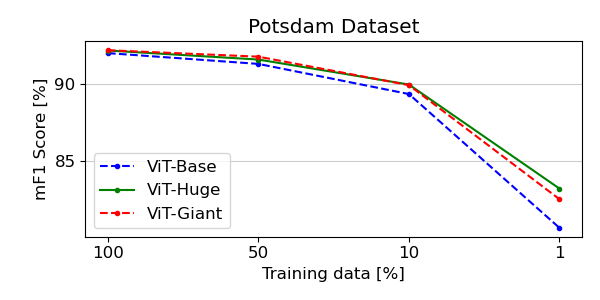}
         \caption{Semantic segmentation (Potsdam)}
         \label{fig:sseg-dataperc}
     \end{subfigure}    
     \begin{subfigure}[b]{0.32\linewidth}
     \centering
        \includegraphics[clip, trim=0 0.2cm 0 0.1cm, width=0.95\linewidth, height=2.5cm]{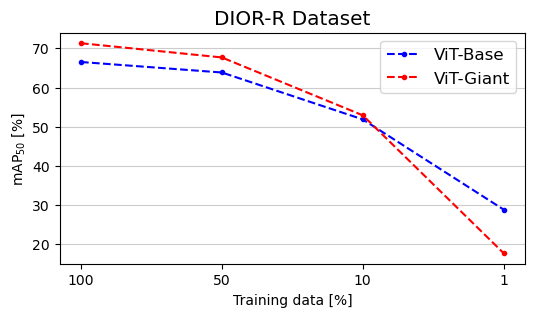}
         \caption{Object detection (DIOR-R)}
         \label{fig:objdet-dataperc}
     \end{subfigure}    
     \caption{Convergence curves for object detection configurations, and finetuning results for object detection and semantic segmentation experiments with varying training set sizes.}  
\end{figure*}

Figure \ref{fig:objdet-dataperc} additionally summarizes performances of our models when using $1$, $10$, and $50\%$ of the total available training data for DIOR-R (object detection) and Potsdam (semantic segmentation) datasets. Here we highlight a lack of reproducibility details on related works, which report maintaining the same FT configuration when performing experiments with different data percentages. Such description is insufficient when training duration and schedules are reported in terms of number of epochs instead of iterations. For our experiments, we opted for maintaining the same number of training iterations for training data budgets considered, since the goal is to assess the sample efficiency of the models, not their sensitivity to training duration (iterations). Our results show consistent benefits of a larger model when $>10\% (1170)$ training images of DIOR-R are used, 
and as low as $1\% (35)$ training images for Potsdam. For DIOR-R, ViT-B values at lower sample budgets are significantly higher than the ones reported by \cite{cha2023billion}, which we conjecture is due to our configuration maintaining the same number of iterations as when finetuning with full dataset.

\begin{table}[htp]
  \caption{Results for finetuning \textsc{OReole-MR} with frozen backbone (Frozen/Not-Frozen)}
  \label{tab:frozen}
  \begin{tabular}{llcc}
    \toprule
    \textbf{Method} & \textbf{Backbone} & \textbf{\makecell{Potsdam \\{}[mF1$\%$]}} & \textbf{\makecell{DIOR-R \\{}[mAP$\%$]}}  \\
    \midrule
    \textsc{OReole-MR} & ViT-B & 88.97/92.01 & 68.33/68.58 \\ 
    \textsc{OReole-MR} & ViT-H & 89.75/92.18 & / \\
    \textsc{OReole-MR} & ViT-G (1B) & 89.62/92.20 & 71.06/71.31\\
  \bottomrule
\end{tabular}
\end{table}

\begin{figure*}[h]
    \centering
    \includegraphics[width=0.8\textwidth]{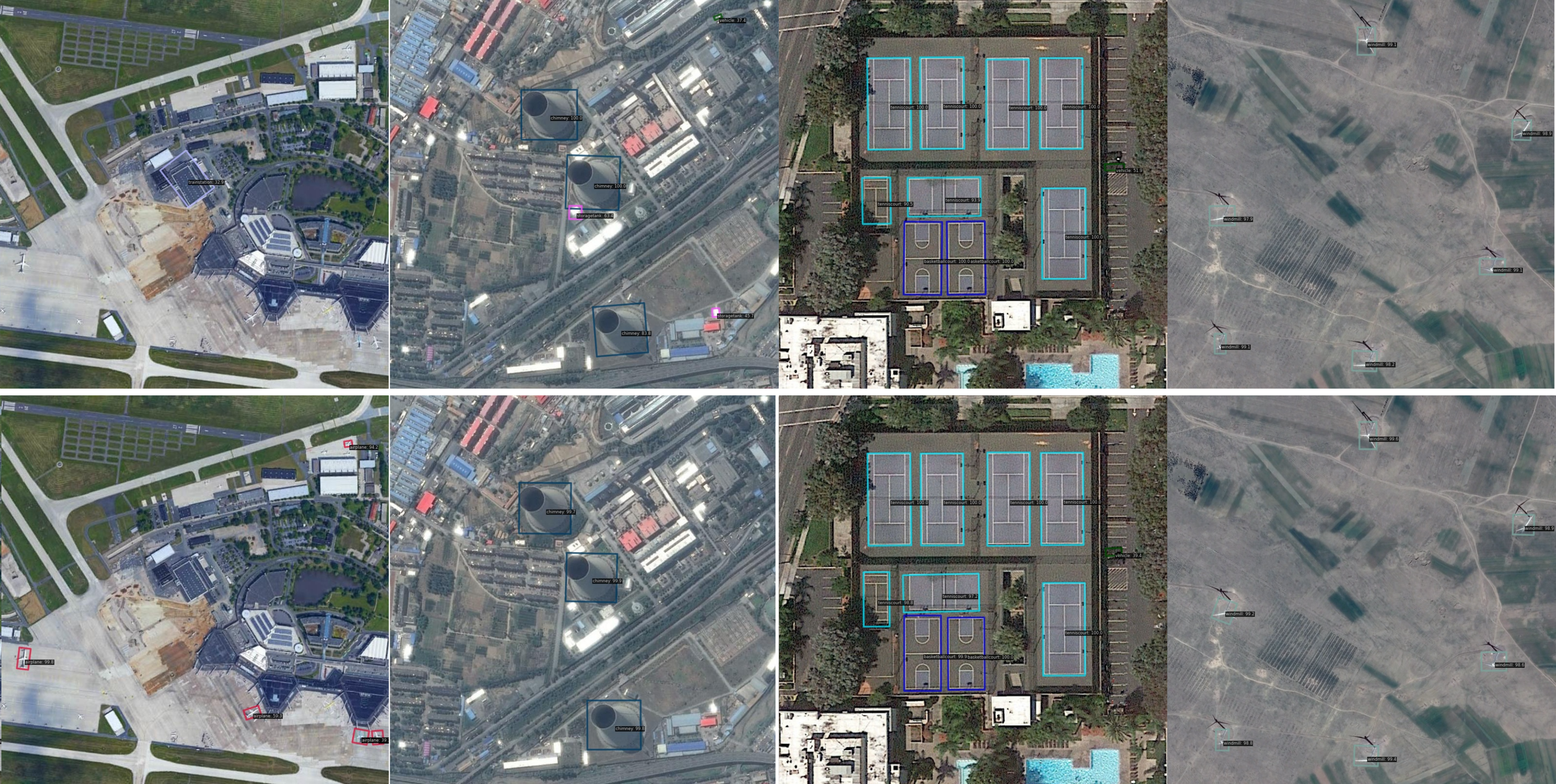}
    \caption{Examples of detections provided by our \textsc{OReole-MR} ViT-B(top) and ViT-G(1B) (bottom) model variants for composing the test set of the DIOR-R dataset.}
    \label{fig:dior_quali}
\end{figure*}

\noindent\textbf{Qualitative examples} Figure \ref{fig:dior_quali} provides qualitative examples comparing outputs provided by ViT-B(top) and ViT-G(1B) (bottom) model variants for composing the test set of the DIOR-R dataset. Noticeable improvements are the detections of airplanes by $ViT-g(1B)$ in the first image, and removal of false positives in the second example from the left. 

\subsection{\textsc{OReole-HR} additional results}
\begin{figure}[h]
    \centering
    \includegraphics[clip, trim=0 0.45cm 0 0.1cm, width=0.5\linewidth]{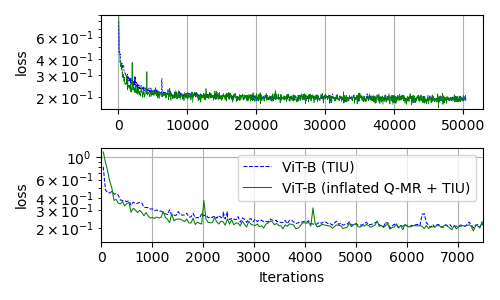}
    \caption{Loss curves for \textsc{OReole-HR} models pretrained from scratch vs with \textsc{OReole-MR} pretrained weights inflated with an extra randomly initialized 4th band.}
    \label{fig:loss_inflated}
\end{figure}

\noindent\textbf{Initialization using inflated OReole-MR pretrained weights} In addition to improved $F1$ results discussed in Section 5, the loss curves in \ref{fig:loss_inflated} show how initialization with pretrained \textsc{OReole-MR} weights inflated with a 4th band yield faster convergence (i.e., better starting point) for MAE pretraining on our 4-band datasets.  

\begin{figure}[h]
    \centering
    \includegraphics[width=0.65\linewidth]{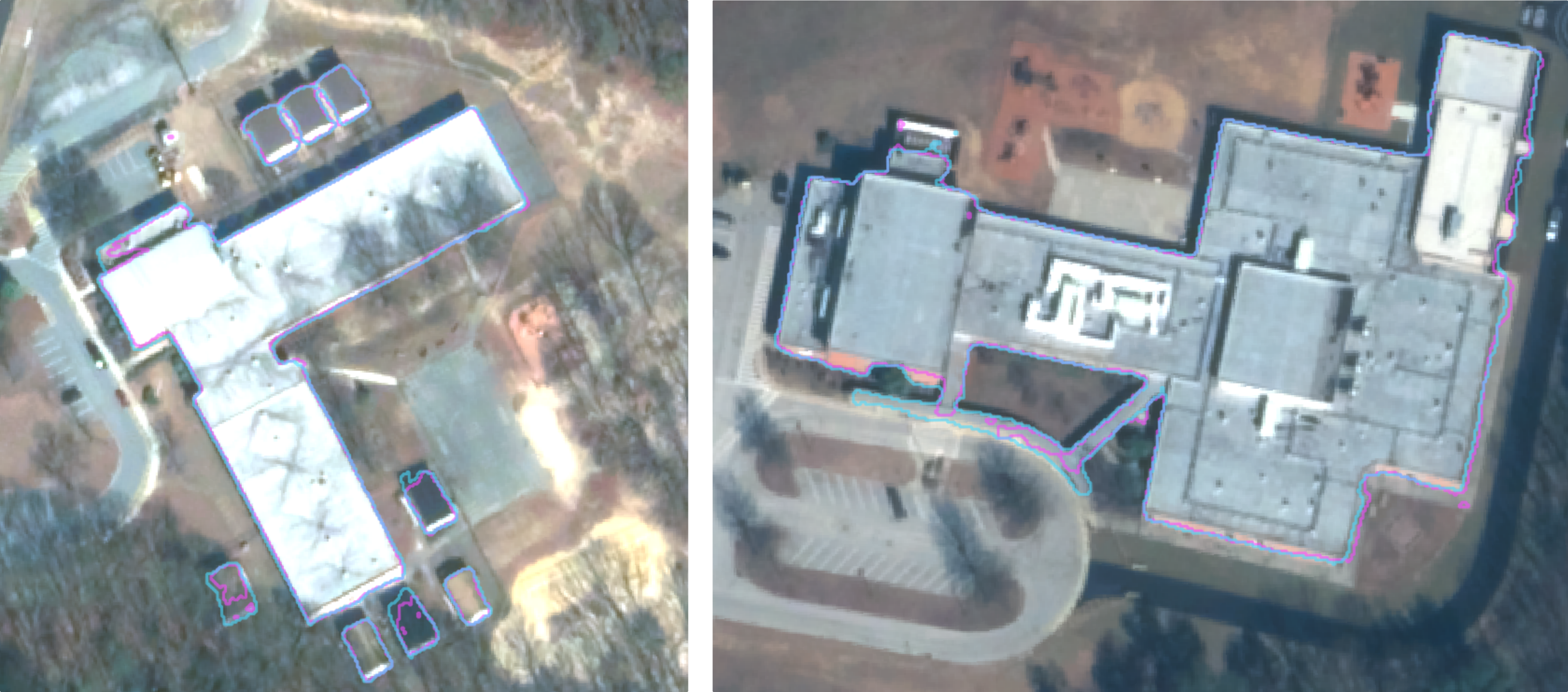}
    \caption{Qualitative example comparing BFE outputs by a ViT-H pretrained in TIU (in blue) compared to a ViT-B pretrained on ORB only (in pink).}
    \label{fig:bfe}
\end{figure}

\noindent\textbf{Qualitative results} Figure \ref{fig:bfe} provides a qualitative example comparing BFE outputs by a ViT-H pretrained in TIU (in blue) compared to a ViT-B pretrained on ORB only (in pink). ViT-H outputs improve adherence to building boundaries in both examples, and retrieving better delineations including in areas of low image contrast (e.g., lower-left building in left-most example).

\end{document}